\documentclass[reprint,superscriptaddress,amsmath,amssymb,aps,pre]{revtex4-1}

\usepackage{lineno,hyperref}
\usepackage{import}
\modulolinenumbers[5]

\usepackage{color,soul}

\usepackage{todonotes}
\usepackage{gensymb}
\usepackage[textwidth=19cm,textheight=24cm]{geometry}
\usepackage{csvsimple}
\usepackage{booktabs, tabularx, wrapfig}
\usepackage{array, ltablex, multirow}
\usepackage{placeins}
\usepackage{graphicx}
\usepackage[table]{colortbl}  % colored table cells
\usepackage[caption=false]{subfig}
\usepackage{prettyref}
\usepackage{color}

\renewcommand{\floatpagefraction}{.8} %fraction beyond which float
\usepackage[nonumberlist,nogroupskip,nomain,acronym]{glossaries}
\glsdisablehyper

\hypersetup{%
  colorlinks=false,% hyperlinks will be black
  linkbordercolor=[rgb]{0.6,0.6,0.6},% hyperlink borders will be red
  pdfborderstyle={/S/U/W 1}% border style will be underline of width 1pt
}

\AtBeginEnvironment{tabular}{\footnotesize}

\newrefformat{tab}{Tab.~\ref{#1}}
\newrefformat{sec}{Section~\ref{#1}}
\newrefformat{fig}{Fig.~\ref{#1}}
\newrefformat{sifig}{App.~Fig.~\ref{#1}}

\newacronym{FCN}{FCN}{fully convolutional network}
\newacronym{JCD}{\text{JCD}}{Jaccard Distance}
\newacronym{DD}{\text{RDD}}{Relative Diameter Deviation}
\newacronym{CD}{\text{RCD}}{Relative Circularity Deviation}
\newacronym{ISF}{\text{ISF}}{Invalid Spheroid Fraction}
\newacronym{ASF}{\text{ASF}}{Ambiguous Spheroid Fraction}

\begin{document} % approx. 3500 words
    % \glsunsetall % reset in section metrics
\title{Image segmentation of treated and untreated tumor spheroids by Fully Convolutional Networks}

\author{Matthias Streller}
\affiliation{DataMedAssist Group, HTW Dresden-University of Applied Sciences, 01069 Dresden, Germany}
\affiliation{Faculty of Informatics/Mathematics, HTW Dresden-University of Applied Sciences, 01069 Dresden, Germany}

\author{So\v{n}a Michl\'ikov\'a}
\affiliation{OncoRay - National Center for Radiation Research in Oncology, Faculty of Medicine and University Hospital Carl Gustav Carus, TUD Dresden University of Technology, Helmholtz-Zentrum Dresden-Rossendorf, Dresden, Germany}
\affiliation{Helmholtz-Zentrum Dresden-Rossendorf, Institute of Radiooncology - OncoRay, Dresden, Germany}

\author{Willy Ciecior}
\affiliation{DataMedAssist Group, HTW Dresden-University of Applied Sciences, 01069 Dresden, Germany}
\affiliation{Faculty of Informatics/Mathematics, HTW Dresden-University of Applied Sciences, 01069 Dresden, Germany}

\author{Katharina L\"onnecke}
\affiliation{DataMedAssist Group, HTW Dresden-University of Applied Sciences, 01069 Dresden, Germany}
\affiliation{Faculty of Informatics/Mathematics, HTW Dresden-University of Applied Sciences, 01069 Dresden, Germany}

\author{Leoni A. Kunz-Schughart}
\affiliation{OncoRay - National Center for Radiation Research in Oncology, Faculty of Medicine and University Hospital Carl Gustav Carus, TUD Dresden University of Technology, Helmholtz-Zentrum Dresden-Rossendorf, Dresden, Germany}
\affiliation{National Center for Tumor Diseases (NCT), NCT/UCC Dresden, Germany}

\author{Steffen
  Lange}
% \email{steffen.lange@tu-dresden.de}
\thanks{These authors contributed equally to this work and share last authorship.}
\affiliation{DataMedAssist Group, HTW Dresden-University of Applied
  Sciences, 01069 Dresden, Germany}
\affiliation{OncoRay - National Center for Radiation Research in Oncology, Faculty of Medicine and University Hospital Carl Gustav Carus, TUD Dresden University of Technology, Helmholtz-Zentrum Dresden-Rossendorf, Dresden, Germany}

% \ead{steffen.lange@tu-dresden.de}
\author{Anja Voss-B\"ohme}
\thanks{These authors contributed equally to this work and share last authorship.}
\affiliation{DataMedAssist Group, HTW Dresden-University of Applied Sciences, 01069 Dresden, Germany}
\affiliation{Faculty of Informatics/Mathematics, HTW Dresden-University of Applied Sciences, 01069 Dresden, Germany}

% \fntext[sharedlastauthor]{These authors contributed equally to this work and share last authorship.}
% \cortext[correspondingauthor]{Corresponding author}

\begin{abstract} % approx. 350 words
\paragraph*{Background and objectives}

Multicellular tumor spheroids (MCTS) are advanced cell culture systems
for assessing the impact of combinatorial radio(chemo)therapy as they
exhibit therapeutically relevant in-vivo-like characteristics from 3D
cell-cell and cell-matrix interactions to radial pathophysiological
gradients. State-of-the-art assays quantify long-term curative
endpoints based on collected brightfield image time series from large
treated spheroid populations per irradiation dose and treatment arm.
This analyses require laborious spheroid segmentation of up to 100.000
images per treatment arm to extract relevant structural information
from the images, e.g., diameter, area, volume and circularity. While
several image analysis algorithms are available for spheroid
segmentation, they all focus on compact MCTS with clearly
distinguishable outer rim throughout growth. However, they often fail
for the common case of treated MCTS, which may partly be detached and
destroyed and are usually obscured by dead cell debris.

\paragraph*{Results}
To address these issues, we successfully train two Fully Convolutional
Networks, UNet and HRNet, and optimize their hyperparameters to
develop an automatic segmentation for both untreated and treated MCTS.
We extensively test the automatic segmentation on larger, independent
data sets and observe high accuracy for most images with Jaccard
indices around 90\%. For cases with lower accuracy, we demonstrate
that the deviation is comparable to the inter-observer variability. We
also test against previously published datasets and spheroid
segmentations.

\paragraph*{Conclusion}
The developed automatic segmentation cannot only be used directly but
also integrated into existing spheroid analysis pipelines and tools.
This facilitates the analysis of 3D spheroid assay experiments and
contributes to the reproducibility and standardization of this
preclinical \emph{in-vitro} model.

\end{abstract}

\maketitle

% \begin{keyword}
% Spheroids, Organoids, Brightfield Microscopy, Segmentation, Deep
% Learning, Fully Convolutional Networks, High-content screening, cancer
% 3D models, cancer therapy
% \end{keyword}

    %\linenumbers

\section{Introduction}
% \glsresetall

One of the most challenging problems in oncology is the development of
therapeutic approaches for tumor growth suppression and eradication,
as well as designing and optimizing treatment protocols. In this
context, three-dimensional (3D) multicellular tumor spheroids (MCTS)
are an advocated pre-clinical, in-vitro model to systematically study
possible means of tumor suppression, assess the curative effect of
combinatorial radio(chemo)therapy, and predict the response of
\emph{in-vivo}
tumors~\cite{FraMicAlaKunVosLan2023,BruZieRivOelHaa2019,BruRivBoxOelHaa2020,RifHeg2017,
  LeeGriHarMcI2016,HirMenDitWesMueKun2010,FriEbnKun2007,KunFreHofEbn2004}.
In contrast to two-dimensional (2D) clonogenic survival assays, which
are known to reflect the therapeutic response of cancer cells in the
tissue comparatively poorly~\cite{Peretal2020}, multicellular
spheroids are reproducible 3D avascular clusters of several thousand
tumor cells without or in advanced settings with stromal cell
compartments mimicking the pathophysiological milieu of tumor
microareas or micrometastases. Due to their more or less radial 3D
histomorphology and structure, they exhibit many characteristic
features affecting tumor growth dynamics, including 3D reciprocal
cell-cell and cell-extracellular-matrix interactions as well as
metabolic gradients of oxygen, nutrients and waste products, which
strongly impair the cells’ proliferative activity and therapy
response~\cite{RifHeg2017,LeeGriHarMcI2016,HirMenDitWesMueKun2010,FriEbnKun2007,KunFreHofEbn2004},
in particular oxygen deficiency (hypoxia), which is associated with
substantial radioresistance~\cite{HowAlp1957, MuzPueAzaAza2015,
  RakEscFurMeiLar2021, LiWisOkoEtal2022}. After a long phase in which
spheroids were used only in specialized laboratories, methodological
advances in serial culturing and live imaging have led to an
exponential spread of this in-vitro model system over the past two
decades~\cite{KunFreHofEbn2004,FriEbnKun2007,Peietal2021}.

While 3D tumor spheroids provide a physiologically more realistic
\emph{in-vitro} framework to study tumor growth dynamics and
therapeutic outcomes, the analysis of the experiments is much more
complex than for traditional 2D cultures. 3D tumor spheroids' dynamics
with and without treatment are most frequently monitored via
microscopy imaging. State-of-the-art long-term spheroid-based assays
assess therapy response from these image time series by classifying
each spheroid within a population as either controlled or relapsed
based on their growth kinetics. By averaging the therapeutic response
over ensembles of spheroids for each treatment dose, the spheroid
control probabilities and spheroid control doses are computed as
analytical endpoints~\cite{CheManLehSorLoeYuDubBauKesStaKun2021,
  CheMicEckWonMenKraMcLKun2021,
  HinIngKaeLoeTemKoeDeuVovStaKun2018,VynBobGarDitStaKun2012},
analogous to the tumor control dose employed in \emph{in-vivo}
radiotherapy experiments with tumor-bearing
mice~\cite{KumLoeGurHerYarEicBauKraJen2021, BauPetKra2005}. Moreover,
time points of relapse are identified to quantify growth recovery in
terms of Kaplan-Meier curves. These metrics depend on the spheroid
type (cell line), the size of the spheroids at the onset of treatment,
and the applied treatment and often require the extraction of growth
curves, i.e., spheroid volume over time, from the image time series of
all individual spheroids.

The associated data analysis poses a significant, interdisciplinary
challenge: A single, typical experimental series of a long-term
spheroid-based assay, i.e., 2-3 cell models at ten different doses
with 3-5 agents and 30 spheroids per treatment arm monitored up to 60
days after treatment, generates up to $10^5$ images. In the past,
several tools were developed for spheroid image
analysis~\cite{HoqWinLovAve2013,IvaParWalAleAshGelGar2014,
  MonFerDucArgDegYou2016,
  RueSchHinDeZWalAreEli2017,MorPalSanTorPapPilSpiGum2017,BouVosTitCruMicFer2018},
including SpheroidSizer~\cite{Cheetal2014},
AnaSP~\cite{Pic2015,PicPeiStePyuTumTazWevTesMarCas2023},
TASI~\cite{HouKonBraMarCoo2018}, SpheroidJ~\cite{Lacetal2021}, and
INSIDIA~\cite{MorPalSanTorPapPilSpiGum2017,PerRosFriPieBarParCiaMorPapSpiPal2022}.
The most crucial part of spheroid image analysis is the identification
of the set of pixels in an image that corresponds to the spheroid, as
this forms the basis for the extraction of spheroid characteristics
like diameter, volume, and circularity as well as other morphologic
features. This identification or classification of pixels in a given
image is denoted semantic image segmentation~\cite{Jad2020} and
represents the greatest challenge in spheroid image analysis. Previous
spheroid segmentations were based on classical segmentation
techniques, including thresholding methods (Yen~\cite{YenChaCha1995},
Otsu~\cite{Ots1979,Pic2015}), watershed algorithm~\cite{RoeMei2000},
shape-based detections (Hough transform algorithms~\cite{DudHar1972},
active contour models/snake ~\cite{CasKimSap1997}), or edge detection
(Canny, Sobel~\cite{Lacetal2021}), and are also available as ImageJ
plugins~\cite{IvaParWalAleAshGelGar2014,MorPalSanTorPapPilSpiGum2017},
Matlab packages~\cite{HouKonBraMarCoo2018,Cheetal2014,Pic2015}, or
dedicated segmentation
programs~\cite{CisOanStuReg2016,VinGowBoxPatZimCouLomMenHarEcc2012,CorCikGus2012}.
Since the characteristics of spheroid images, including the size,
shape, and texture of the actual spheroid, vary with cell line,
treatment, and microscopy method (e.g.,
brightfield~\cite{SadKarWae2017,Lacetal2021,MorPalSanTorPapPilSpiGum2017,PerRosFriPieBarParCiaMorPapSpiPal2022,PicPeiStePyuTumTazWevTesMarCas2023},
fluorescence~\cite{Lacetal2021,CelRizBlaMasGliPogHas2014,PicTesZanBev2017,MorPalSanTorPapPilSpiGum2017,BouVosTitCruMicFer2018,PerRosFriPieBarParCiaMorPapSpiPal2022},
differential, interference
contrast~\cite{NgoYanMaoNguNgKuoTsaSawTu2023}), most of these
approaches and tools are specialized for specific experimental
conditions or even selected
microscopes~\cite{CelRizBlaMasGliPogHas2014,PicTesZanBev2017} and
often fail to generalize to arbitrary conditions. In recent years,
data-based methods, especially deep-learning models, have been
increasingly employed for spheroid segmentation to tackle this issue
of
generalization~\cite{SadKarWae2017,Zaoetal2021,Lacetal2021,ZhuCheWanTanHeQinYanJinYuJinLiKet2022,NgoYanMaoNguNgKuoTsaSawTu2023,AksKatAbeSheBesBurBigAdaMonGhe2023,PicPeiStePyuTumTazWevTesMarCas2023,GarDomHerMatPas2024}.
Deep-learning segmentation also allows more complex analysis, e.g.,
identification of time-lapse migratory
patterns~\cite{NgoYanMaoNguNgKuoTsaSawTu2023} or distinction of the
spheroids' core and invasive
edge~\cite{PerRosFriPieBarParCiaMorPapSpiPal2022}.

\begin{figure*}
    \centering
    \subfloat[Treated - 10 Gy]{\includegraphics[width=0.32\linewidth]{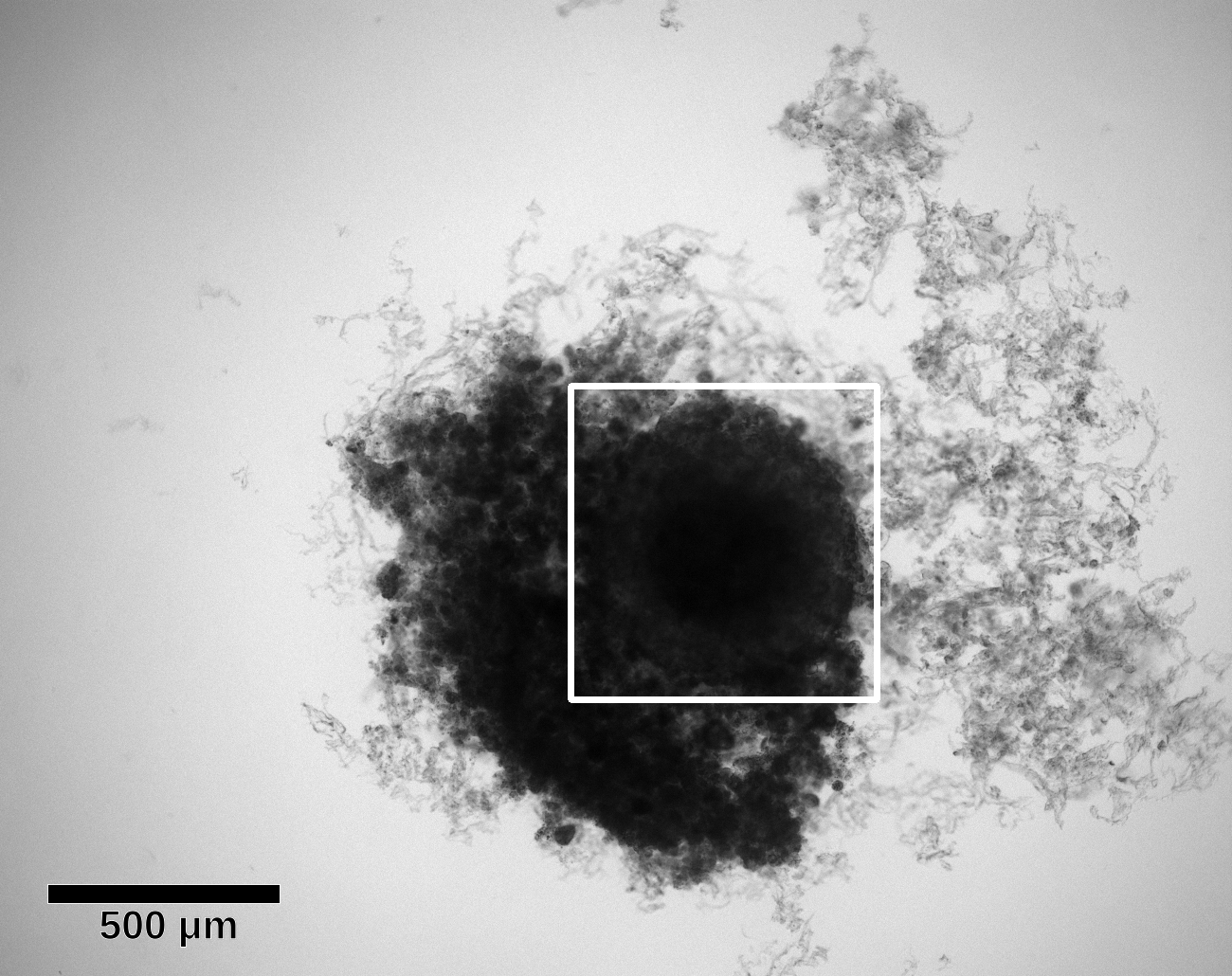}}
    \quad
    \subfloat[Zoom of white box in (a)]{\includegraphics[width=0.32\linewidth]{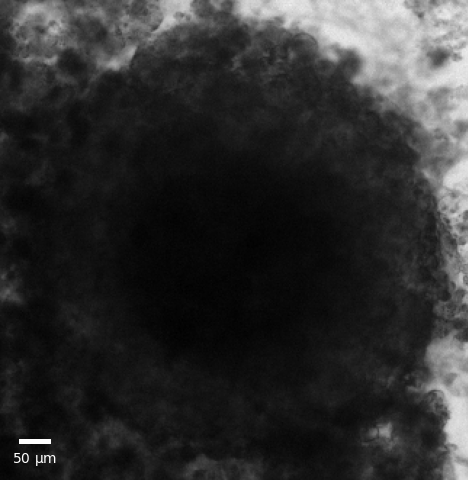}}
    \quad
    \subfloat[Segmentation\label{fig:exmpl_hard}]{\includegraphics[width=0.32\linewidth]{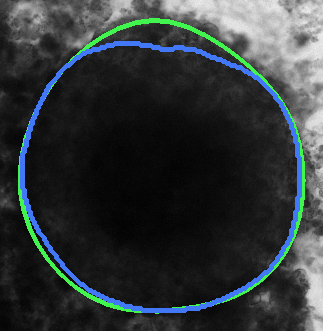}}

    \subfloat[Treated - 10 Gy]{\includegraphics[width=0.32\linewidth]{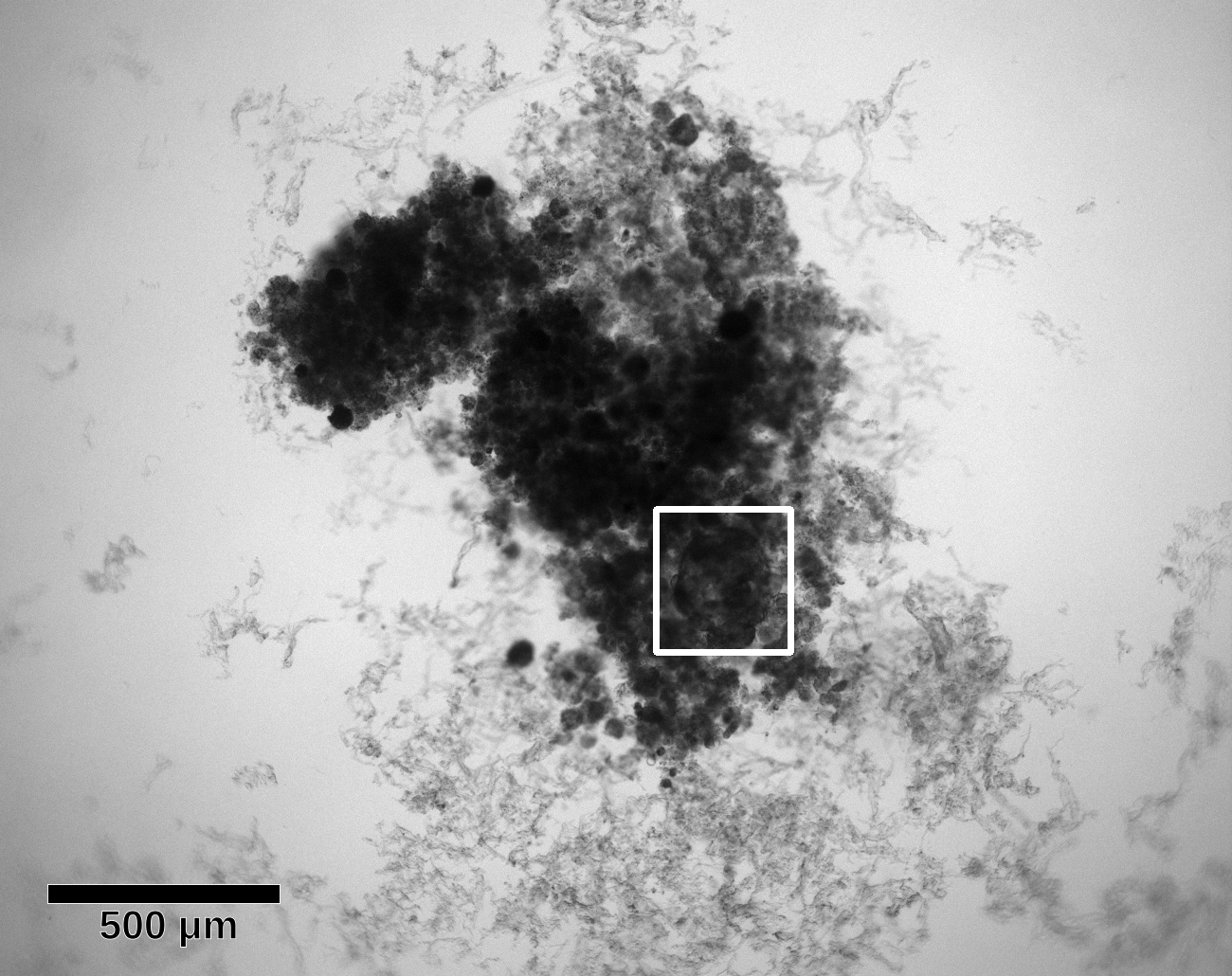}}
    \quad
    \subfloat[Zoom of white box in (d)]{\includegraphics[width=0.32\linewidth]{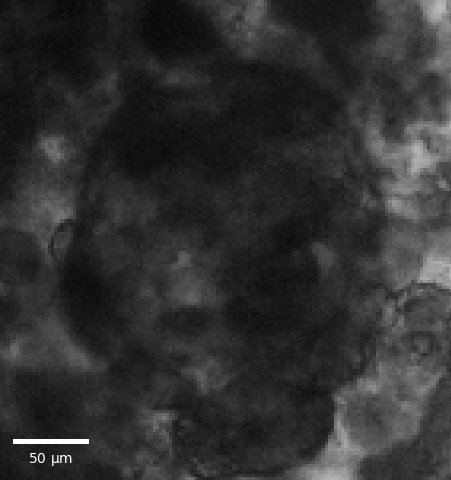}}
    \quad
    \subfloat[Segmentation\label{fig:exmpl_easy}]{\includegraphics[width=0.32\linewidth]{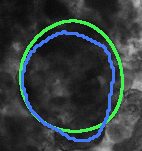}}
    \caption{Representative examples of images for automatic
      segmentation with the optimized U-Net (blue) compared to the
      manual segmentation (green) for 3D tumor spheroids after
      treatment. The overlap with the manual segmentation is excellent
      for standard size and larger spheroids obscured by cell debris
      (top row) and sufficient for small, heavily obscured spheroids
      (bottom row). Displayed are (a,d) the original images, which are
      also the input for the U-Net, (b,d) magnified image details
      around the spheroids as indicated by the white box in (a,d) for
      visibility, and (c,e) corresponding contours from the
      segmentations. The metrics for evaluation of the automatic
      segmentation are: the top row - \gls{JCD} $ = 7.6\%$, \gls{DD}
      $= 3.7\%$ and \gls{CD} $= 1.6\%$; bottom row -
      $\gls{JCD} = 22.7\%$, $\gls{DD}= 7.5\%$ and $\gls{CD}= 14.7\%$,
      see text for details.}
\label{fig:exmpl-seg}
\end{figure*}

However, these segmentation and analysis tools have primarily been
developed for experimental conditions where the resulting images are
relatively clean with well-visible, unobscured MCTS. In contrast,
radiotherapy, one of the most common cancer treatments, regularly
causes spheroids to shed dead cells or even to detach completely.
Consequently, dead cell debris often obscures the remaining shrunk
MCTS and the 3D cultures putatively regrowing from surviving viable
cells after detachment, see \prettyref{fig:exmpl-seg} for two
examples. This debris can cover a much larger domain than the actual
spheroid and can be locally even thicker and thus darker, which makes
segmentation difficult even for human experts. Both classical
segmentation techniques and previous deep-learning models typically
fail in these cases and require manual adjustments. We illustrate this
challenge of spheroid segmentation by applying the four most recently
proposed deep-learning
models~\cite{Lacetal2021,AksKatAbeSheBesBurBigAdaMonGhe2023,PicPeiStePyuTumTazWevTesMarCas2023}
to images with and without severe cell debris, see
\prettyref{sifig:model-comparison} for representative examples and
\prettyref{sitab:our-data} for statistical results.

We train deep-learning models to segment spheroid images after
radiotherapy based on annotated images from a previous experiment
assessing the radioresponse of human head-and-neck squamous cell
carcinoma (HNSCC). After systematic optimization of hyperparameters
and pre-processing for two selected network architectures, U-Net and
HRNet, the automatic segmentation exhibits high accuracy for images of
both treated and untreated spheroids. We further validate the
automatic segmentation with the optimized U-Net on larger, independent
data sets of two cell types of head-and-neck cancer exposed to both
radiotherapy and hyperthermia. For the majority of images, we find
excellent overlap between manual and automatic segmentation, even in
the case of small, heavily obscured spheroids, see
\prettyref{fig:exmpl-seg}. In particular, we compare segmentations of
different biological experts and find that imprecisions of the
automatic segmentation are comparable to variations across different
humans. The optimized automatic segmentation is provided in a minimal
tool with a graphical user interface.

\section{Materials}

The original data is compiled from several data sets of time images of
FaDu cell line spheroids, one of the two head-and-neck cancer cell
types from the previously published study on combinatorial
radioresponse in human head-and-neck squamous cell carcinoma spheroid
models~\cite{CheMicEckWonMenKraMcLKun2021}: Spheroid populations were
irradiated with X-rays at single doses of either 2.5 Gy, 5.0 Gy, 7.5
Gy or 10.0 Gy, with each dose being represented approximately equally
in the data set. Spheroids were imaged for up to 60 days after
treatment with brightfield microscopy, i.e., at different times after
treatment and with potentially different lighting conditions. 1095
spheroids were manually segmented by a biological expert (human H1) to
train and test the \glspl{FCN}. We use the train-validation-test-split
technique for validation during training, which is appropriate for the
large datasets and commonly used to train spheroid segmentation
models~\cite{Lacetal2021,AksKatAbeSheBesBurBigAdaMonGhe2023,PicPeiStePyuTumTazWevTesMarCas2023}.
We split the annotated data into 3 separate data sets: 883 images for
training, 108 for validation, and 104 for testing (Hold-out test set).
Images from a single spheroid are exclusively assigned to one of the
groups to rule out unwanted correlations and data leakage between
training, validation, and testing data sets. For further testing,
another set of 6574 images of head-and-neck cancer (FaDu and SAS)
spheroids from the same study~\cite{CheMicEckWonMenKraMcLKun2021},
treated with different combinations and doses of X-ray irradiation and
hyperthermia, is manually segmented by several independent biological
experts (humans H2-H5). Note that while the automatic segmentation is
performed on individual images, biological experts always take
precedent and subsequent images of a spheroid into account for manual
segmentation.

The original images were taken with a Zeiss Axiovert 200M at a
resolution of 1300 × 1030 pixels, representing $2.04\ \mu$m/pixel with
16-bit gray-scale per pixel~\cite{CheMicEckWonMenKraMcLKun2021}.
Images are converted to 8-bit images to ensure compatibility with the
employed libraries FastAi~\cite{HowGug2020} and
SemTorch~\cite{Cas2020}. While this conversion, in principle, reduces
the contrast level, the effect is negligible as only a small fraction
of the 16-bit range is utilized during imaging, e.g., the mean
gray-scale value in the images is $1300\pm130$, and values are
rescaled such that smallest gray-scale value of the original image
becomes 0 and the biggest value becomes 255.

\begin{figure*}
    \centering
    \subfloat[Valid spheroid
    segmentation\label{fig:metric_valid}]{\includegraphics[width=0.32\linewidth]{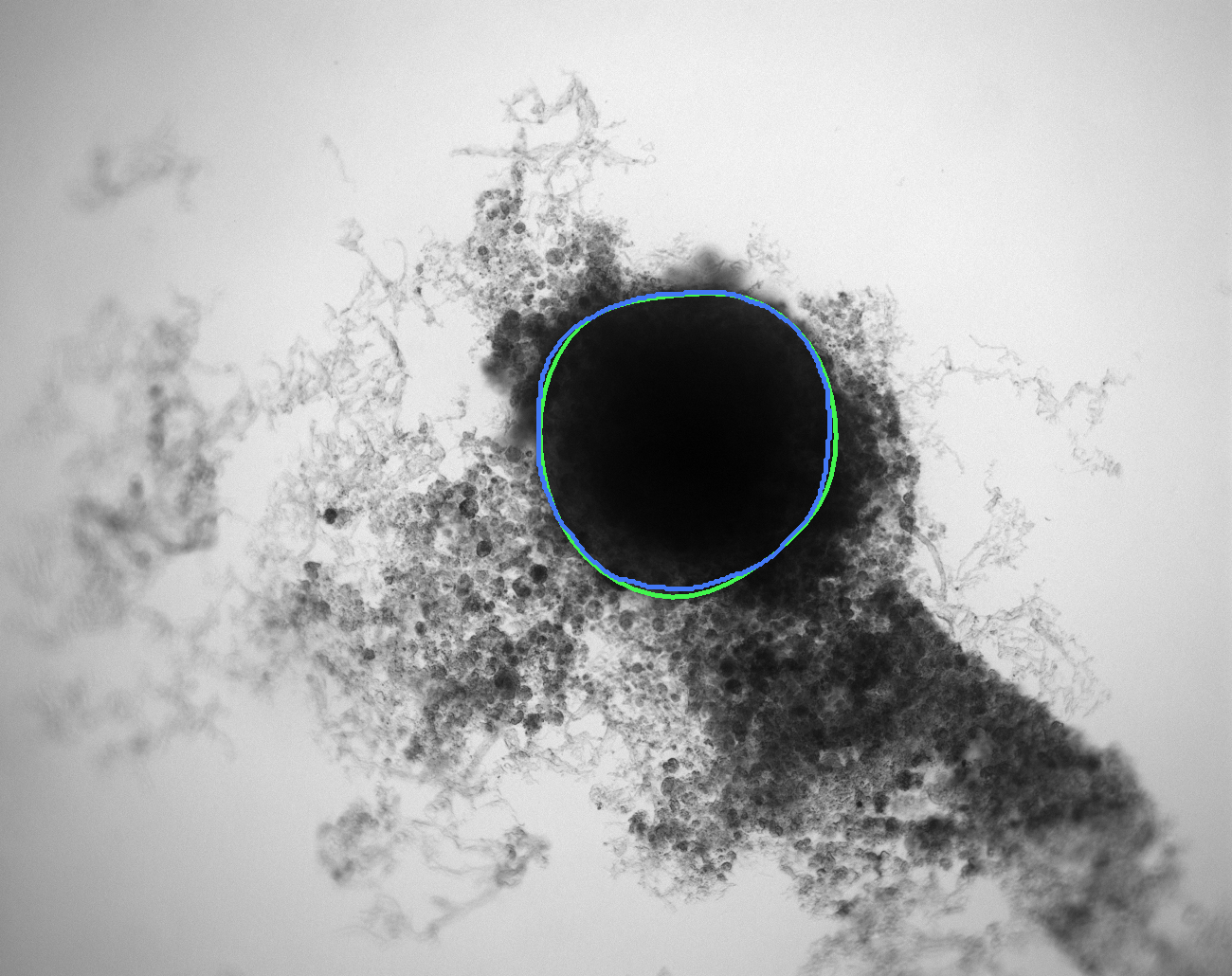}}
    \quad
    \subfloat[Ambiguous spheroid segmentation\label{fig:metric_ambiguous}]{\includegraphics[width=0.32\linewidth]{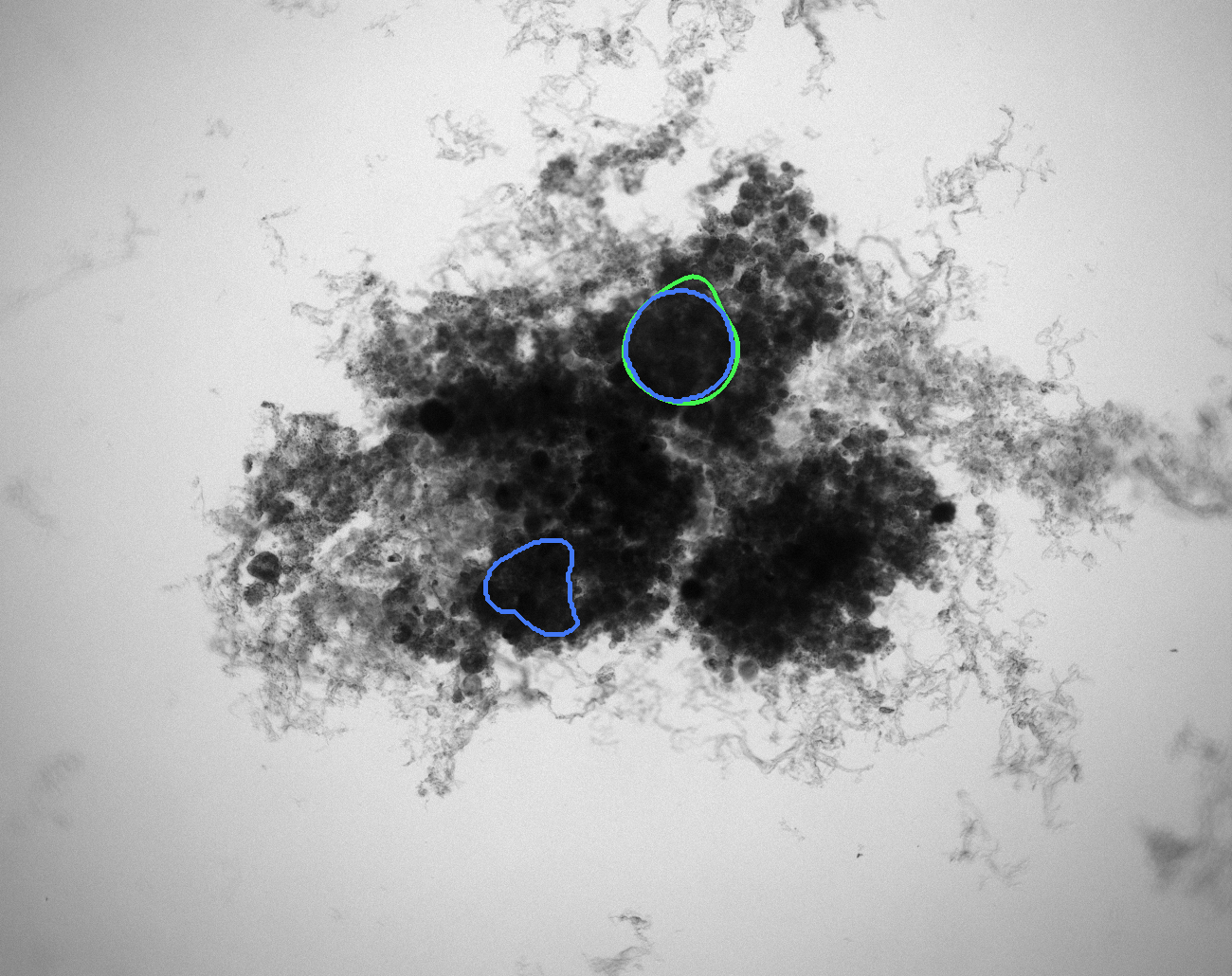}}
    \quad
    \subfloat[Invalid spheroid segmentation\label{fig:metric_invalid}]{\includegraphics[width=0.32\linewidth]{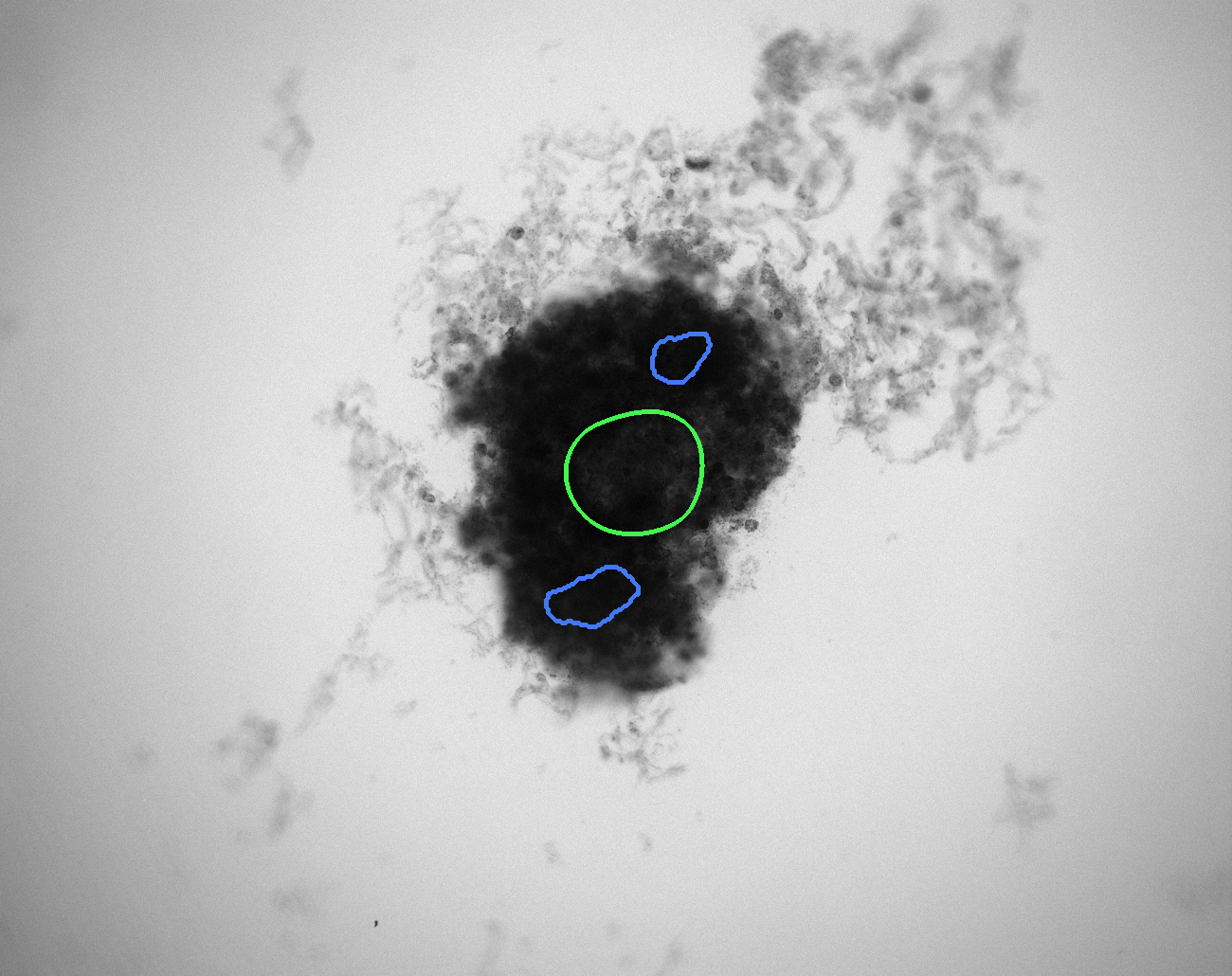}}
    \label{fig:metric_samples}
    \caption{Exemplary evaluation of the automatic segmentation (blue)
      with respect to the manual one (green) for (a) standard case and
      (b),(c) rare artifacts, see text for details: (a) Correctly
      segmented spheroid: no contribution to \gls{ISF} or \gls{ASF};
      \gls{JCD}, \gls{DD} and \gls{CD} are well defined. (b) Excessive
      spheroids are detected beyond the actual spheroid: no
      contribution to \gls{ISF}, one count added to \gls{ASF};
      \gls{JCD}, \gls{DD} and \gls{CD} are well defined and computed
      for the larger, upper spheroid. (c) No overlap between automatic
      and manual segmentation: one count added to \gls{ISF}, no
      contribution to \gls{ASF}; \gls{JCD}, \gls{DD} and \gls{CD} are
      set to one.}
\end{figure*}

As deep-learning models, we use \glspl{FCN}, a particular group of
convolutional neural networks intended for semantic image
segmentation~\cite{LonSheDar2015}, in particular the
U-Net~\cite{RonFisBro2015} and the HRNet \cite{Wanetal2021}. The U-Net
is one of the most established \gls{FCN} frameworks. The HRNet has a
special architecture compared to other typical \gls{FCN} models and
has already been successfully employed in the case of clean images
with clearly visible spheroids~\cite{Lacetal2021}. The pipeline to
train the models employs FastAi~\cite{HowGug2020} for the U-Net and
its backbones and SemTorch~\cite{Cas2020} for the HRNet. Before each
training, a suitable learning rate is picked using LR
Range~\cite{Smi2018}. The training follows the 1-cycle policy: the
learning rate is cycled once between two bounds, allowing fast
convergence~\cite{Smi2018}. During training we use validation-based
early stopping, where the \gls{JCD} is the underlying metric. The
models are trained on a GPU NVIDIA GeForce RTX 3080 with a mini-batch
size of 2. For training on images with higher resolution or to compare
some specific backbones, online learning is used instead of
mini-batches to avoid memory overflow, i.e., the batch size, the
amount of data used in each sub-epoch weight change, is reduced to
one.

\section{Methods}
\glsresetall
\glsunset{FCN}

The segmentation is evaluated via several metrics
(\prettyref{sec:metrics}: \gls{JCD}, \gls{DD}, \gls{CD}, \gls{ASF},
\gls{ISF}) additional to the standard accuracy quantified by the
Jaccard coefficient of the whole data set, which is used during
training. These metrics are computed for individual images to report
not only their mean values but also their deviations, and thus
reliability, across the test data set. These metrics are used to
optimize the hyperparameters, \prettyref{sec:hyperparameter}, and data
augmentation, \prettyref{sec:augmentation}, of the selected \gls{FCN}
models U-Net and HRNet for maximal accuracy.

\subsection{Evaluation metrics}\label{sec:metrics}
\glsresetall
\glsunset{FCN}

Several metrics are employed to asses the accuracy of the
segmentation. The most important one is the \gls{JCD}, which measures
the relative difference between two sets of pixel $P$ and $T$,

\begin{equation}\label{eq:JCD}
    \gls{JCD} = 1 - \text{IoU} = 1 - \frac{|P \cap T|}{|P \cup T|}.
\end{equation}

with the automatically segmented (predicted) pixel set $P$ from the
\gls{FCN} and the (target) pixel set $T$ of the manually segmented
spheroid. Note that in the literature often the opposite metric
Intersection over Union $\text{IoU} = 1 - \gls{JCD}$ also called
Jaccard index, or Jaccard similarity coefficient, is used. The
\gls{JCD} takes values between 0 (automatic segmentation perfectly
overlaps with the manual one) and 1 (no intersection between the two
segmentations). The \gls{JCD} can be understood as the relative area
error of the segmentation, i.e., in the example of
\prettyref{fig:metric_valid} $\gls{JCD} = 0.042$ is obtained, which
means $4.2\%$ error in the segmented area. A selection of sample
images with different \gls{JCD}s is displayed in
Figs.~\ref{sifig:samplesJCD10}-\ref{sifig:samplesJCD30} to give an
optical reference for this metric. From these sample images and in
accordance with the biological experts, it may be concluded that a
\gls{JCD} below $0.2$ is justifiable and such a deviation is observed
between segmentations of different humans, see also
\prettyref{sec:further-validation}.

\begin{figure*}
    \centering
    \subfloat[Probability
    heatmap]{\includegraphics[width=0.32\linewidth,trim={0.25cm 0.25cm
        0.25cm 0.25cm},clip]{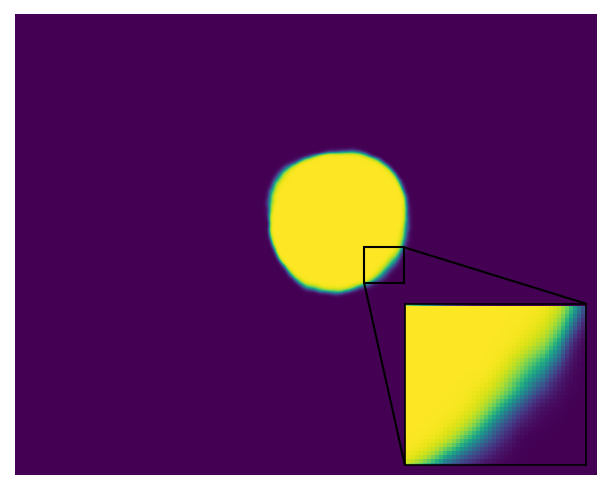}}
    \quad
    \subfloat[Segmentation map]{\includegraphics[width=0.32\linewidth,trim={0.25cm 0.25cm
        0.25cm 0.25cm},clip]{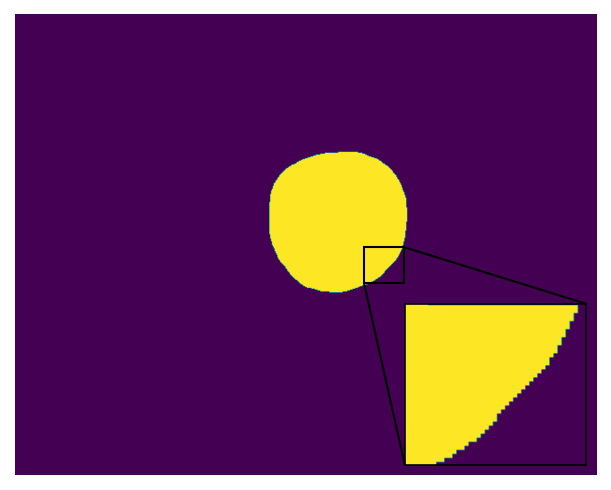}}
    \quad
    \subfloat[Final polygonal chain]{\includegraphics[width=0.32\linewidth,trim={0.25cm 0.25cm
        0.25cm 0.25cm},clip]{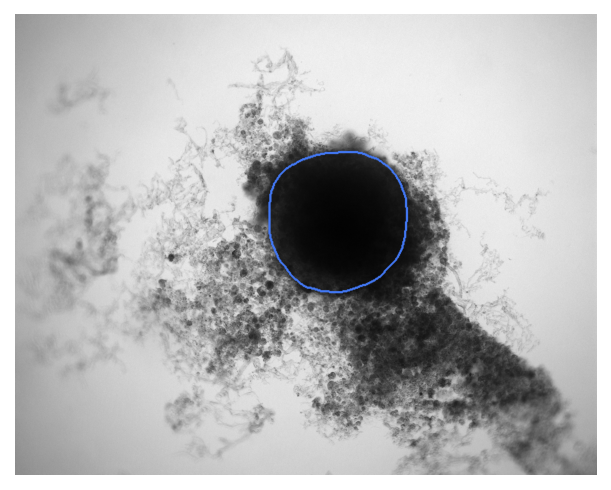}}
    \caption{Post-processing steps for the output of the \gls{FCN}
      model to transform probability heatmap to spheroid contour: (a)
      Probability heatmap as output of the \gls{FCN} model. Each pixel
      takes probability values between 0 (black) and 1 (white)
      predicting the target (spheroid). Note that due to the steep
      gradient the gradual change from black to white is hard to see.
      (b) By using a threshold of 0.5, the pixels are classified into
      outside spheroid (black) and inside spheroid (white). (c)
      Contour of the spheroid border extracted as a polygonal chain
      (blue line displayed on original image).}
    \label{fig:postprocess}
\end{figure*}

It is possible that the spheroid is correctly segmented, but
additionally another non-existent spheroid is detected, see
\prettyref{fig:metric_ambiguous}, where additionally to the manually
segmented spheroid in the top right another contour in the bottom left
is suggested by the automatic segmentation. In this case, the
\gls{JCD} would be below 1 but differ from 0. The fraction of these
cases among all images are additionally denoted \gls{ASF}.
Furthermore, a \gls{JCD} of 1 can mean that the spheroid is found in
the wrong place or the spheroid is not found at all, see
\prettyref{fig:metric_invalid}. The fraction of these cases is denoted
\gls{ISF}.

In practice, only the average diameter $d_T = 2 \sqrt{|T|/\pi}$ of the
segmented spheroid is extracted to estimate the three-dimensional
volume $\pi d_T^3/6$ of the spheroid under the assumption of a
spherical shape. Additionally, the circularity $4 \pi |T|/L^2$ of the
spheroid is computed from its area $|T|$ and perimeter $L$ to assess
the validity of this assumption. Note that this is just a common
assumption in the field, although even a circular projection does not
imply a spherical shape. Thus, we also assess the automatic
segmentation by the resulting error of these two values and measure
the \gls{DD} and \gls{CD} between the segmentations

\begin{equation}
  \gls{DD}, \gls{CD} = \frac{|c_P-c_T|}{c_T}
\end{equation}

where $c_P$ is either mean diameter or circularity based on the
automatic segmentation and $c_T$ is the corresponding characteristic
calculated based on the manual segmentation. In the rare case that
several spheroids are detected in an image, the largest one is chosen
for computation of the \gls{JCD}. This strategy is in accordance with
experimental standard practice for single spheroid-based assays with
curative analytical endpoints, as the largest regrowing spheroid is
sufficient to detect relapse after treatment. Note that if there were
several spheroids present and of interest, also all of them could be
segmented. In the case of $\gls{JCD} = 1$, which corresponds to an
invalid spheroid, \gls{DD} and \gls{CD} are also set to 1. The
combination of the \gls{JCD}, \gls{ASF}, \gls{ISF}, \gls{DD}, and
\gls{CD} quantify the accuracy of the segmentation with the \gls{FCN},
where always a value closer to zero means higher agreement between the
segmentations.

The spheroid's pixel set and perimeter are required to calculate the
metrics. Thus, probability heatmaps, created by the Softmax function
in the last layer of the \gls{FCN}, are transformed into binary images
and polygonal chains as illustrated in \prettyref{fig:postprocess}.
First, the binary image is created. For this purpose, every pixel with
a probability value higher or equal to 50\% will be counted as a
spheroid pixel and set to 1. Values of the other pixels are set to 0.
The coordinates from the outer border of the emerged shape in the
binary image can be extracted by the algorithm proposed
in~\cite{SuzBe1985}, thereby creating the final polygonal chain.

\subsection{Hyperparameter optimization \label{sec:hyperparameter}}

Several hyperparameters and pre-processing methods can be adjusted to
improve the accuracy of the selected \gls{FCN} models. We successively
optimize the most relevant parameters, starting from the ones with the
most significant expected impact on accuracy to the one with the
lowest impact, i.e., \emph{backbone}, \emph{transfer learning},
\emph{data augmentation}, \emph{input resolution}, \emph{loss
  function}, and \emph{optimizer}: The \emph{backbone} defines the
exact architecture of the \gls{FCN} model and is accordingly the most
important hyperparameter. For \emph{transfer learning}, the
ImageNet~\cite{Imagenet} data set is utilized. The \emph{data
  augmentation} techniques are described in
\prettyref{sec:augmentation}. Furthermore, the impact of changing the
\emph{input resolution} is investigated. Afterward, the \emph{loss
  function} for quantifying the \gls{FCN}'s error is optimized,
particularly by comparing distribution-based loss and a region-based
loss. We pick the Dice loss for region-based loss and the
Cross-Entropy loss as a distribution-based loss function. We also test
the Focal loss, which is a variation of Cross-Entropy and suitable for
imbalanced class scenarios~\cite{Jad2020}. Finally, two different
optimizers for minimizing the loss function are evaluated. One is
Adam~\cite{KinBa2017}, a very common optimizer, and the other one is a
combination of some more modern optimizers, RAdam~\cite{Liuetal2021}
and Lookahead~\cite{Zhaetal2019}. For the backbone, the optimization
is initiated with a default setting, i.e., transfer learning, no data
augmentation, input resolution half of original image, Cross-Entropy
loss function, and Adam optimizer. These initial parameters are then
individually optimized in consecutive order.

\subsection{Data Augmentation}\label{sec:augmentation}

Data augmentation is a method to generate additional training data via
transformations of the original data set. Since this increases the
total amount of training data, data augmentation can improve
performance and avoid overfitting~\cite{Busetal2020}. In our case,
every original training image is transformed once, doubling the size
of the original training data set. For each image in the training data
set, one of three transformations is picked randomly using
Albumentations~\cite{Busetal2020}, i.e., vertical flip, horizontal
flip, or rotation by $180\degree$. These transformations are valid,
since the spheroids do not have a specific orientation within an
image. Note that each of these transformation conserves the original
rectangular resolution of the image. This avoids extrapolating pixels
on the border of the image, which would be necessary for arbitrary
rotation angles and which can lead to additional image artifacts the
\gls{FCN} model would have to be trained for.

\section{Results}

\subsection{Hyperparameter optimization}

For each hyperparameter set and pre-processing method a model is
trained and the performance of these trained models is compared for
the validation data set based on the proposed evaluation metrics from
\prettyref{sec:metrics}. For \gls{JCD}, \gls{DD}, and \gls{CD} not
only the average value but also the standard deviation over the whole
validation data set is reported to estimate reliability of the
segmentation. For the \gls{ISF} and the \gls{ASF} the standard error
of the mean of the corresponding binomial distribution is reported.
All results of the optimization are listed in the supplemental
material (Figs.~\ref{sifig:backbone_unet}-\ref{sifig:optimizer_hrnet})
and briefly discussed in the following.

\subsubsection{Backbones}

For the U-Net, the ResNet~34 and the VGG~19-bn architecture perform
best, see \prettyref{sifig:backbone_unet}, with a \gls{JCD} three
times smaller than the worst performing backbones, like AlexNet or
SqueezeNet1.0. The two ResNet backbones stand out with a very low
\gls{ASF}. Note that the accuracy of the VGG architectures is
improving with increasing number of used layers. The ResNet~34 is the
optimal backbone as it not only performs best in virtually every
metric, but also requires less computational time than the VGG~19-bn.
Note that a previous study on deep-learning models for spheroid
segmentation observed a higher performance for ResNet~18 compared to
VGG~19 and ResNet~50~\cite{PicPeiStePyuTumTazWevTesMarCas2023}. For
the HRNet, the backbones differ in the number of kernels used and the
accuracy increases with the number of kernels, see
\prettyref{sifig:backbone_hrnet}. The W48 backbone is preferred, as
for the majority of metrics it performs best and has the lowest
standard deviations.

\subsubsection{Data augmentation and transfer learning}

The combination of data augmentation and transfer learning leads to
the best performance for both U-Net and HRNet, see
\prettyref{sifig:extension_unet} and
\prettyref{sifig:extension_hrnet}. While this is in principle
expected, the improvement is substantial as the \gls{JCD} is more than
halved when both transfer learning and data augmentation is
introduced. Note that for the U-Net, transfer learning improves
accuracy more than data augmentation, while the opposite is true for
the HRNet.

\subsubsection{Resolution}

Using input images with half the resolution $650 \times 515$ of the
original images leads to the best performance, see
\prettyref{sifig:resolution_unet} and
\prettyref{sifig:resolution_hrnet}. This input resolution is already
used to optimize the previous parameters. Higher resolution (3/4, 1)
reduces accuracy, presumably because the increase in details makes
generalization of the appearance of spheroids more difficult, and the
receptive fields of the kernels are getting smaller relative to the
whole image. Lower resolution (1/4) also reduces accuracy, presumably
due to the loss of relevant information. Half resolution displays the
best accuracy and lowest standard deviations across the validation
data set.

\subsubsection{Loss}

We find the Dice loss to be the optimal loss function for the U-Net,
see \prettyref{sifig:loss_unet}. Note that Dice is considered the
default loss parameter when using the U-Net for cell segmentation,
i.e., binary classification concerned with the accuracy of the edge.
However, the advantage is much smaller than for previous
hyperparameters. It does not suggest a general advantage of
region-based loss for the segmentation of spheroids by the U-Net. For
the HRNet, there is a more significant difference between the
region-based loss and the distribution-based loss, see
\prettyref{sifig:loss_hrnet}. The Focal loss and the Cross-Entropy
loss perform better than the Dice loss - especially, the standard
deviations differ by a factor of up to three. One reason for this
behavior may be that the Dice loss provides less detailed feedback
during training than the distribution-based losses. Cross-Entropy loss
and Focal loss achieve nearly the same results, and we choose the
default Cross-Entropy loss.

\subsubsection{Optimizer}

In case of the U-Net the difference between the tested optimizers is
marginal, see \prettyref{sifig:optimizer_unet}. We choose the slighly
better performing RAdam combined with Lookahead over the default
optimizer Adam. In contrast, the accuracy of the HRNet is much worse
for the combined optimizer, see \prettyref{sifig:optimizer_hrnet}. The
Adam optimizer is optimal with evaluation metrics 1.5 times better
than for the RAdam combined with Lookahead.

\begin{table}[!h]
  \centering
  \caption{(a) Evaluation of the segmentation with the optimized U-Net
    and HRNet models on the test data set shows higher accuracy of the
    U-Net. Note that all metrics can range between zero and one, where
    lower values mean higher accuracy and zero means perfect agreement
    with the manual segmentation. (b) Evaluation of the optimized
    segmentations on images of untreated (not irradiated) spheroids
    shows high accuracy also for the case of clean images without cell
    debris. The optimized U-Net and HRNet achieve almost equal
    accuracy. (c) The optimal hyperparameter configurations.}
    \label{tab:testset}
    \subfloat[Segmentation test data set]{\label{tab:testsetall}
        \begin{tabular}{l|rrrrr}
            \hline
            Model & \gls{JCD} & \gls{DD} & \gls{CD} & \gls{ISF} & \gls{ASF} \\
            \hline
            \hline
            U-Net & \textbf{0.062} & \textbf{0.020} & 0.040 & \textbf{0.000} & \textbf{0.010} \\
            $\pm$ & 0.060 & 0.026 & 0.034 & 0.000 & 0.010 \\
            \hline
            HRNet & 0.072 & 0.028 & \textbf{0.036} & \textbf{0.000} & \textbf{0.010} \\
            $\pm$ & 0.100 & 0.080 & 0.029 & 0.000 & 0.010 \\
            \hline
        \end{tabular}
      }

      \subfloat[Segmentation images without cell debris]{\label{tab:compacttestset}
    \begin{tabular}{l|rrrrr}
        \hline
        Model & \gls{JCD} & \gls{DD} & \gls{CD} & \gls{ISF} & \gls{ASF} \\
        \hline
        \hline
        U-Net & \textbf{0.028} & \textbf{0.008} & 0.034 & \textbf{0.000} & \textbf{0.000} \\
        $\pm$ & 0.010 & 0.008 & 0.032 & 0.000 & \\
        \hline
        HRNet & 0.029 & \textbf{0.008} & \textbf{0.033} & \textbf{0.000} & \textbf{0.000} \\
        $\pm$ & 0.011 & 0.007 & 0.026 & 0.000 & \\
        \hline
    \end{tabular}}

      \subfloat[Optimized parameters]{\label{tab:optimal-parameters}
    \begin{tabular}{l|cc}
        \hline
        Parameter & U-Net & HRNet \\
        \hline
      \hline
      Backbone & ResNet 34 & W48 \\
      Transfer learning & \multicolumn{2}{c}{Yes} \\
      Data augmentation & \multicolumn{2}{c}{Yes} \\
      Resize factor & \multicolumn{2}{c}{1/2} \\
      Loss function & Dice & Cross-Entropy \\
      Optimizer & RAdam \& Lookahead & Adam \\
        \hline
    \end{tabular}}
\end{table}

\subsection{U-Net vs HRNet}

The automatic segmentations with the final optimized U-Net and HRNet
model are applied to the test data set. While this test data set is
not taken into account during training of the models, the
segmentations display virtually the same high accuracy on these new
images as for the validation data set, see see
\prettyref{tab:testsetall} and \prettyref{sifig:SItestset} cf.
\prettyref{sifig:optimizer_unet} and
\prettyref{sifig:optimizer_hrnet}. Mostly, the standard deviations of
the metrics are higher in the test data set, though the \gls{JCD}
remains even with this deviation well below $20 \%$. As for the
validation data set, the U-Net performs slightly better than the HRNet
on the test data. The \gls{JCD} and \gls{DD} for the HRNet are 0.01
higher, and their standard deviation is 0.04 higher. Two
representative examples of the automatic segmentation by the U-Net are
shown in \prettyref{fig:exmpl-seg}: The overlap with the manual
segmentation is excellent for standard size and larger spheroids
obscured by cell debris, see \prettyref{fig:exmpl-seg} top row, and
sufficient for small, heavily obscured spheroids, see bottom row. In
the latter case, the model does not segment the spheroid boundary
accurately but detects the spheroid at the correct position \gls{JCD}
$= 23 \%$ with a comparable size $\gls{DD} = 7.5 \%$ within the much
larger and often darker cloud of surrounding cell debris. In such
difficult cases, similar variations are observed upon manual
segmentation by different biological experts.

While the models are mainly trained on images with treated spheroids
which are obscured by cell debris, they also accurately segment
untreated spheroids with clear boundary and clean background, see
\prettyref{tab:compacttestset}: On a test set of 100 images of
untreated spheroids, the accuracy of both models is nearly the same
with a \gls{JCD} below $3\%$. In comparison, a recent deep-learning
approach for spheroids with clear boundary and clean background with
focus on generalizability to different experimental conditions and
microscopes, achieved a mean \gls{JCD} of $8\%$ with a standard
deviation of $12\%$ for bright-field microscopic
images~\cite{Lacetal2021}. This suggests that the standard case of
clean images without cell debris is included in our training of images
with heavily obscured spheroids.

The final U-Net model has a size of 158 MB and the final HRNet model a
size of 251 MB. For both models, it takes about 1.8 seconds to segment
one image on the CPU (Intel(R) Core(TM) i7-4770). When the
segmentation is performed serially on the GPU (NVIDIA GeForce RTX
3080) the U-Net needs only 0.03 seconds per image and the HRNet 0.08
seconds. As computation time for both models is comparable, the U-Net
is chosen for the automatic segmentation due to its slightly higher
accuracy in our setting. Note that in a recently published
deep-learning approach, the comparison of HRNet and U-Net implied that
HRNet achieved the highest accuracy~\cite{Lacetal2021}.

\begin{figure}[!ht]
  \centering
  \subfloat[]{\label{fig:JCD-diameter}\hspace*{-0.5cm}\includegraphics{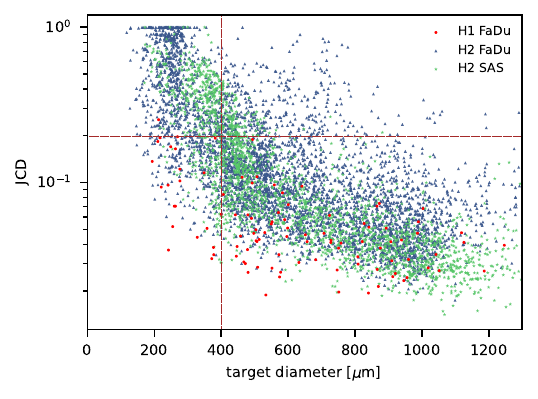}}

  \subfloat[]{\label{fig:dr-diameter}\hspace*{-0.5cm}\includegraphics{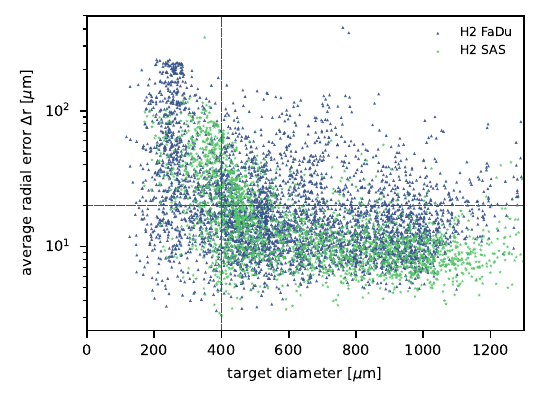}}
  \caption{Validation of the automatic segmentation with the optimized
    U-Net on larger, independent data sets shows high accuracy for the
    majority of cases. (a) \gls{JCD} and (b) average radial error
    $\Delta r$ over diameter of the manually segmented (target)
    spheroid $d_T$ for 6574 images of FaDu (blue triangles) and SAS
    (green stars) spheroids treated with different combinations and
    doses of X-ray irradiation and
    hyperthermia~\cite{CheMicEckWonMenKraMcLKun2021}. Manual
    segmentation is performed by a second biological expert (human H2,
    blue triangles and green stars) independently from the manual
    segmentation (human H1, red dots) of the training, validation and
    test data sets. (Results for $104$ images of test data set is
    displayed as red dots for comparison.) Note that the segmentation
    is developed only based on images of FaDu spheroids. The majority
    of deviations are small ($\gls{JCD}<0.2$, $\Delta r < 20\ \mu$m,
    red horizontal lines as guide to the eye), average (median) values
    are $\gls{JCD} = 0.17 (0.09) \pm 0.2$,
    $\Delta r = 25 (15) \pm 32\ \mu$m for the whole data set and
    $\gls{JCD} = 0.1 (0.07) \pm 0.09$,
    $\Delta r = 18 (13) \pm 17 \ \mu$m for spheroids larger
    $d_T\geq400\ \mu$m (red vertical lines as guide to the eye) than
    the initial, standard size of spheroids. Larger imprecisions for
    smaller spheroids are due to biologically difficult, unclear, or
    ambiguous cases, see text.}
\label{fig:validation}
\end{figure}

\subsection{Further independent validation \label{sec:further-validation}}

\begin{figure*}
  \centering
  \includegraphics[width=\linewidth]{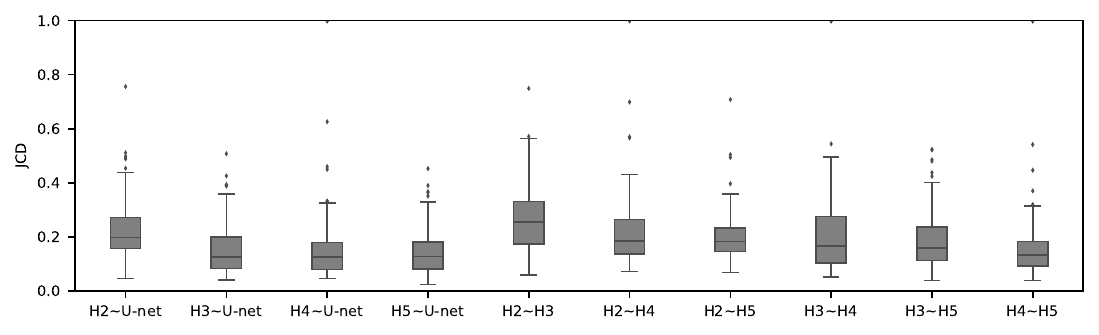}
  \caption{For treated spheroids smaller than the initial, standard
    size before treatment ($d\leq 400\ \mu$m) deviations from the
    manual segmentation are not higher than variations across
    segmentations by different humans, suggesting that the
    segmentation of images with small spheroids surrounded by heavy
    debris is often difficult or ambiguous: Compared are segmentations
    from the optimized U-Net and 4 independent human experts (H2-H5)
    for the same 101 images, which are randomly selected from the pool
    of small spheroids of the extended Hold-out test data set, see
    \prettyref{fig:validation}. The Friedman test score of all
    \gls{JCD}s 217.2 ($p \ll 0.001$) indicates significant differences
    among the pairwise segmentation deviations. The order according to
    the average JCD is (from low to high values) H5$\sim$U-Net,
    H3$\sim$U-Net, H3$\sim$H5, H2$\sim$H5 $<$ H4$\sim$H5,
    H4$\sim$U-Net, H2$\sim$U-Net, H3$\sim$H4, H2$\sim$H4, H2$\sim$H3,
    where the $<$ indicates a significant ($p<0.005$) difference
    between the sets of \gls{JCD}s according to a Dunn-Bonferroni
    pairwise post-hoc test.}
\label{fig:ai-human}
\end{figure*}

In addition to the standard check with the Hold-out test data set, we
systematically validate the automatic segmentation with the optimized
U-Net on larger, independent Hold-out test data sets of two cell types
of head-and-neck cancer, see \prettyref{fig:validation}. This data set
contains 6574 images of FaDu or SAS spheroids treated with different
combinations and doses of X-ray irradiation and
hyperthermia~\cite{CheMicEckWonMenKraMcLKun2021}. These images were
manually segmented by another biological expert (human H2) independent
from the original training/validation/test-data set used to develop
the automatic segmentation (human H1). For the majority of images, we
find excellent overlap between this manual and the automatic
segmentation, quantified by a \gls{JCD} around 0.1, see
\prettyref{fig:JCD-diameter}. Larger deviations are mostly observed
for images with smaller spheroids, i.e., below the size of standard
spheroids for treatment (diameter $370-400\ \mu$m according
to~\cite{CheManLehSorLoeYuDubBauKesStaKun2021,CheMicEckWonMenKraMcLKun2021,HinIngKaeLoeTemKoeDeuVovStaKun2018}).
To make the evaluation intuitive for the biological experts, who are
less familiar with the \gls{JCD} defined in Eq.~(\ref{eq:JCD}), we
additionally introduce a measure of the average radial error
$\Delta r$ based on the automatically segmented (predicted) domain $P$
and the manually segmented (target) domain $T$
\begin{equation} \label{eq:dr}
  \Delta r = \sqrt{\pi^{-1}\left(|P \cup T|-|P \cap T|\right) + \frac{d_T^2}{4}}-\frac{d_T}{2}
\end{equation}
with the average (target) diameter $d_T$ of the manually segmented
domain. The error $\Delta r$ quantifies the thickness a circular layer
with the size of the mismatched area
$|P \cup T|-|P \cap T| = |P\setminus T| + |T\setminus P|$ around a
circle with the target area $|T|$ would have, see
\prettyref{sifig:ARE} for an illustration. Since additional
$P\setminus T$ and missing areas $T\setminus P$ do not compensate each
other in Eq.~(\ref{eq:dr}), the error $\Delta r$ is considerably
larger than the radial error implied by the relative difference
between target and predicted diameter \gls{DD}, see
\prettyref{sifig:RDD}. The average radial error $\Delta r$ reported in
\prettyref{fig:dr-diameter} assesses the segmentation analogous to the
\gls{JCD} with majority of errors $\Delta r<20\ \mu$m and larger
deviations mostly for spheroids with diameter clearly smaller than
before treatment ($d_T\leq 370-400\ \mu$m). For reference, note that
the resolution of the images is $2.04 \ \mu\text{m}/\text{px}$, i.e.,
a deviation of $20\ \mu\text{m}$ corresponds to $10$px ($<1$\%
deviation with respect to the original 1300 $\times$ 1030 image) or
$5$px in the half-resolution image for application of the U-Net.

Only a few images with larger spheroids exhibit \gls{JCD} bigger than
$0.2$. Most of these are ambiguous cases, with two spheroids attached
to each other, which are inconsistently recognized as either one or
two spheroids even by the human, see examples with
$\gls{JCD}= 0.28-0.34$ and $0.42$ in \prettyref{sifig:samplesJCD30}.
While the training data set did not contain such double-spheroid
cases, the optimized U-Net often segments them correctly, see examples
with $\gls{JCD}=0.08, 0.12$ in Figs.~\ref{sifig:samplesJCD10}
and~\ref{sifig:samplesJCD20}, respectively. Apart from this particular
case, most images with larger deviations between the segmentations
refers to smaller spheroids with surrounding cell debris. Note that
smaller spheroids without cell debris, i.e., untreated or shortly
after treatment, do not exhibit such deviations but are accurately
segmented. To thoroughly examine these larger deviations in more
detail, three biological experts independently segmented 101 randomly
selected images with spheroid diameter $d_T\leq 400\ \mu$m. For these
101 images the \glspl{JCD} between the humans and the U-Net and
between different humans are compared in \prettyref{fig:ai-human}. We
find that on average the discrepancies between humans and U-Net are
comparable to the variations across segmentations from different
humans. This implies that these images represent biologically
difficult, unclear, or ambiguous cases. It suggests that the larger
\gls{JCD} observed for some smaller spheroids rather reflects this
uncertainty and not a low performance of the U-Net. Accordingly, the
evaluation in \prettyref{fig:validation} refers to a worse-case
scenario, as for each spheroid, the biological expert intentionally
segmented all images over time, while in practice, segmentation is
only required and performed for a fraction of these images, in
particular for clearly distinguishable and larger spheroids.

Finally, we also test the automatic segmentation with the optimized
U-Net on eight published test data sets from previous deep-learning
approaches~\cite{Lacetal2021,IvaParWalAleAshGelGar2014,PicPeiStePyuTumTazWevTesMarCas2023},
see \prettyref{sitab:their-data}. Note that the total of 496 images
originates from different conditions, including brightfield and
fluorescence microscopy, RGB and 16-bit gray-scale, and different
microscopes, magnifications, image resolutions, and cell models, but
do not contain significant cell debris. While our automatic
segmentation performs well on roughly half of the data sets, sometimes
surpassing the original model corresponding to the data set, two types
of images turn out problematic: (i) images with ambiguous ground truth
and (ii) images on which the spheroids appears semi-transparent, with
individual cells being visible throughout the spheroid. However,
classical segmentation techniques work sufficiently well for both
types of images (i) and (ii), making the use of deep-learning
approaches in these cases unnecessary. In detail, images of type (i)
contain spheroids with peculiar intensity profiles, i.e., with a
compact sphere-like core surrounded by a flat patch. The corresponding
published ground truth just defines the outer boundary of these
patches as the boundary of the spheroid, while our automatic
segmentation restricts the spheroid to the compact core. Either choice
may be unreasonable depending on the goal of the analysis. The outer
boundary of the patches allows computation of the projected area of
the spheroid while ignoring the considerable variations in thickness
encoded in the intensity across the spheroid. Accordingly, it is not
suited to estimate the spheroid volume necessary to evaluate volume
growth. Presumably, the compact core contains most of the
three-dimensional organized cells and cellular volume. It is the main
component exhibiting metabolic gradients affecting therapy
response~\cite{RifHeg2017,LeeGriHarMcI2016,HirMenDitWesMueKun2010,FriEbnKun2007,KunFreHofEbn2004},
but certainly, it underestimates the total cell volume. Segmentation
of such incompletely formed spheroids is debatable, even biologically,
as the formation of proliferative and metabolic gradients remains
ambiguous and may depend on the question pursued with the experiment.
For images of type (ii), the apparent visibility of individual cells
throughout the spheroid may be due to its small size or the chosen
microscopy method.

\section{Discussion}

We develop an automatic segmentation for images of 3D tumor spheroids
both with and without (radio)therapy. We systematically validate the
automatic segmentation on larger, independent Hold-out test data sets
of two cell types of head-and-neck cancer spheroid types, including
combinations with hyperthermia treatment. For most images, we find
excellent overlap between manual and automatic segmentation. These
include clearly discernable spheroids and the previously neglected
cases of spheroids critically obscured by cell debris. For images
showing poor overlap of the segmentations, we demonstrate that this
error is comparable to the variations between segmentations from
different biological experts (inter-observer variability). This
suggests that considerable deviations between automatic and manual
segmentations do not necessarily reflect a low performance of the
former but rather a general uncertainty or ambiguity in spheroid
identification.

While the accuracy of (spheroid) segmentations is usually only
quantified in terms of the \gls{JCD} or Jaccard index (IoU), we choose
to additionally report the corresponding implications on the spheroid
diameter derived from the segmentation by $\Delta r$ and \gls{DD}.
This makes the evaluation more intuitive for biological experts, as
the spheroid diameter is the central metric for analysis. For
instance, the spheroid volume required for growth curve documentation
and growth delay is estimated from this diameter, which is connected
to the projected area of the spheroid. We do not explicitly translate
the evaluation to the corresponding spheroid area and volume. However,
this may be estimated from basic scaling arguments, i.e., area and
volume scale square and cubic with the diameter. The circularity of
the spheroid, as quality control for its desired spherical shape, is
taken into account by the \gls{CD}. All of these metrics will converge
to their correct value, when the \gls{JCD} goes to 0 (or Jaccard index
to 1).

In fact, it should be pointed out that the imprecisions of the
segmentation have a smaller effect on the quantification of therapy
response than suggested by the reported \gls{JCD}: As most informative
parameters in long-term spheroid-based assays are the average
diameters, volumes, and circularities of the spheroids. Thus, the
relevant metric to estimate the error of the automatic analysis is not
the \gls{JCD} but the \gls{DD} and \gls{CD}, which are considerably
smaller as even an imprecise segmentation can result in the correct
size or shape of the spheroid. Furthermore, higher deviations are
primarily observed for for treated 3D cultures with diameters below
the initial diameter of standard spheroids $d_T= 370-400 \ \mu$m
before treatment. These are usually images of spheroids in the final
steps of detachment or before cell re-aggregation, growth recovery,
and spheroid relapse. In experimental practice, such cases are
typically not segmented as they are less relevant for the analytical
endpoints. For instance, computation of growth delay requires accuracy
of the spheroid segmentation immediately before treatment (for images
without obscuring cell debris) and at large spheroid sizes
$d_T>600 \ \mu$m, while the images in between have no impact on the
growth delay. Moreover, the validation in \prettyref{fig:validation}
is based on images from the time series of relapsed and controlled
spheroids. However, in practical routine, segmentations of complete
time series are only performed for growth curves and growth delay
assessment in spheroid populations at 100 \% growth recovery. The
average \gls{JCD} is substantially smaller if the validation is
restricted to these cases.

It is important to note that the initial automatic segmentation is
driven by images of only one HNSCC spheroid model (FaDu) with and
without treatment. However, we show that it works equally well for
spheroids from another cell type (SAS), although these SAS spheroids
display a different peripheral shape during regrowth after treatment.
Beyond the systematic validation highlighted herein, the automatic
segmentation has been continuously tested and applied by several
biological/biomedical experts and researchers for over a year during
their ongoing experiments and to retrospectively reanalyze MCTS from
earlier studies. So far, it has been reported that resegmentation is
only necessary in a small fraction of cases, mainly due to optical
artifacts, like out-of-focus images. Overall, the segmentation has in
the meantime been successfully applied to numerous untreated
multicellular spheroid types of different tumor entities and cell
lines, respectively (FaDu, SAS - head and neck; Panc-02(mouse),
Panc-1, PaTu 8902 - pancreas; HCT-116, HT29 - colon; A549, NCI-H23,
NCI-H460 - lung; BT474 - breast; LNCap, DU145 - prostate; U87-MG,
U138-MG, U251-MG - brain/GBM; Hek293 - kidney). These spheroids ranged
between $200 - 1000 \ \mu$m in diameter and images were taken at
different magnifications and resolutions (e.g., $1300 \times 1030$,
$1388\times 1040$ $1920 \times 1440$, $1920 \times 1216$ and
$1.6 - 2.6 \ \mu$m/pixel) as single or Z-stack images at diverse
microscopic devices (Axiovert 200M, AxioObserver Z1 - both from Zeiss;
BioTek Cytation 5 Cell Imaging Multimode Reader - Agilent). Many more
spheroid types are in the pipeline for implementation. The developed
automatic segmentation has also already been applied to contour the
images of selected spheroid types, e.g., FaDu, SAS, Panc-02 or DU145,
after various treatments such as radiotherapy (X-ray, proton),
hyperthermia, chemotherapy and combinatorial
treatment.

We also test the automatic segmentation on eight published test data
sets from previous deep-learning
approaches~\cite{Lacetal2021,IvaParWalAleAshGelGar2014,PicPeiStePyuTumTazWevTesMarCas2023}.
We observe a high accuracy, except for special cases, which, however,
turn out to be well segmentable by classical techniques. While the
automatic segmentation has been continuously tested for over a year on
different cell lines, microscopes, and treatments, it does not
necessarily generalize to arbitrary experimental conditions and
imaging, which will be the focus of future improvements. However, this
problem of domain shift, which can always arise when a model is
applied to data sets with a different data distribution than the
training data, is also an ongoing challenge for deep-learning models
focused on images with clear spheroids and clean
background~\cite{GarDomHerMatPas2024}. In contrast, the focus of this
study is the inclusion of the case of treated spheroids surrounded by
severe cell debris, which is frequent after treatment but often
neglected in automatic segmentation.

Note that the automatic segmentation's accuracy and reliability may
not be exclusive to the finally chosen network architecture and
hyperparameters. Indeed, we find that two very different network
architectures, U-Net and HRNet, achieve very similar performance,
which is consistent with previously reported optimizations of
deep-learning models
~\cite[Tab.~2]{Lacetal2021}~\cite[Tab.~3]{PicPeiStePyuTumTazWevTesMarCas2023}.
The segmentation's accuracy also seems relatively insensitive to the
choice of several hyperparameters, e.g., loss function, optimizer
function, and, to some extent, the resize factor. Instead, it is
plausible, that applying challenging training data, i.e., images with
extensive, severe debris, is crucial for the performance of the
resulting automatic segmentation. The network trained primarily on
such images also works on images with clearly visible spheroids.
Hence, the data sets compiled and annotated for this work are also
publicly provided to support future improvements of spheroid
segmentations. This data set can facilitate the development of more
individual, more customized models, e.g., using the recently
introduced nnU-Net tool~\cite{IseJaeKohPetMai2021}.

The automatic segmentation can be incorporated into existing or future
tools for spheroid analysis or medical image analysis as basis for
further feature extraction including perimeter, complexity, and
multiparametric analysis~\cite{ZhuCheWanTanHeQinYanJinYuJinLiKet2022}.
The segmentation provides a basis for the development of an automatic
classification of spheroid image time series into control and relapse
and could support machine learning methods to forecast tumor spheroid
fate early. To make the deep-learning model available, we provide a
minimal tool with graphical user interface. The ONNX
Runtime~\cite{Ord2021} is used to compile the model and to take it
into production.

\section{Availability of source code and requirements (optional, if code is present)}

\begin{itemize}
  \itemsep0em 
\item Project name: Spheroidsegdedeb (Spheroid segmentation despite debris)
\item Project home page:
  \href{https://igit.informatik.htw-dresden.de/aagef650/spheroidsegdedeb}{\url{https://igit.informatik.htw-dresden.de/aagef650/spheroidsegdedeb}}
\item Operating system(s): Platform independent
\item Programming language: Python >3.8.10
\item Other requirements: Additionally download the model from
  \href{https://wwwpub.zih.tu-dresden.de/~s6079592/SEGMODEL.onnx}{\url{https://wwwpub.zih.tu-dresden.de/~s6079592/SEGMODEL.onnx}}.
  Requirements and usage are documented in the provided README and
  some example images for testing are included in the corresponding
  folder.
\item License: GNU GPL
\item BioTool ID: spheroidsegdedeb
\item SciCrunch ID: SpheroidSegDeDeb, RRID:SCR\_026409
\item SpheroidSegDeDeb is also archived in Software
  Heritage~\cite{snapshot}.
\end{itemize}

\section{Data availability}

Training, validation, and test data used for this work are
available~\cite{bioimage} including data on clean spheroids and the
extended dataset as well as the code to train and evaluate the model
(see above repo training\_scripts/). All additional supporting data
are available in the GigaScience repository, GigaDB~\cite{gigadb},
i.e. all networks optimized for different hyperparameters in the .pth
format. DOME-ML annotations are available in DOME
registry~\cite{dome}.

\section{Declarations}

\subsection{List of abbreviations}
\glsresetall \gls{ASF}, \gls{FCN}, \gls{ISF}, \gls{JCD}, Multicellular
tumor spheroids (MCTS), \gls{CD}, \gls{DD}, Supplemental material (SM)

\subsection{Competing Interests}

The authors declare that they have no competing interests.

\subsection{Funding}

The authors acknowledge that this research has been co-financed by the
EU, the European Social Fund (ESF), and by tax funds on the basis of
the budget passed by the Saxon state parliament (project SAB-Nr.
100382145) and the Bundesministerium f\"ur Bildung und Forschung (BMBF
16dkwn001a/b). The publication of this article is funded by the Open
Access Publication Fund of Hochschule für Technik und Wirtschaft
Dresden – University of Applied Sciences. The funders had no role in
study design, data collection and analysis, decision to publish, or
preparation of the manuscript.

\subsection{Author's Contributions}

Conceptualization, A.V.-B., L.A.K.-S., and S.L.; methodology, A.V.-B.,
S.L., S.M., L.A.K.-S., and M.S..; software, M.S., K.L., and S.L.;
validation, W.C., K.L., S.M., and S.L.; formal analysis, M.S.,
A.V.-B., and S.L.; investigation, M.S., K.L., and S.L.; resources,
L.A.K.-S., and A.V.-B.; data curation, M.S., W.C., K.L., and S.L.;
writing -- original draft preparation, M.S., A.V.-B. and S.L.; writing
-- review and editing, A.V.-B., W.C., S.L., S.M., and L.A.K.-S.;
visualization, M.S., W.C., and S.L.; supervision, L.A.K.-S., A.V.-B.
and S.L.; project administration, L.A.K.-S. and A.V.-B.; funding
acquisition, L.A.K.-S. and A.V.-B. All authors have read and agreed to
the published version of the manuscript.

\section{Acknowledgements}

We are grateful to M.~Wondrak, S.~Al-Jamei, N.~El-Refai, R.~Joseph,
and S.~L.~Prieto for sharing their expertise on tumor spheroid
contouring.

% \bibliographystyle{cpg_unsrt_title}
% \bibliography{\pathtorepo bibtex/abbrevs,\pathtorepo bibtex/datamedassist_refs,additional_refs} % up to 50 refs

\setcounter{figure}{0}
\appendix
\section{Supplemental figures}
\renewcommand{\floatpagefraction}{0.1}
\renewcommand{\thetable}{A.\arabic{table}}
\renewcommand{\thefigure}{A.\arabic{figure}}
\renewcommand{\theequation}{A.\arabic{equation}}
\renewcommand{\thesection}{A.\arabic{section}}

\begin{figure*}[!h]
  \centering
      \subfloat[Untreated spheroid w/o cell debris]{\includegraphics[width=0.49\linewidth]{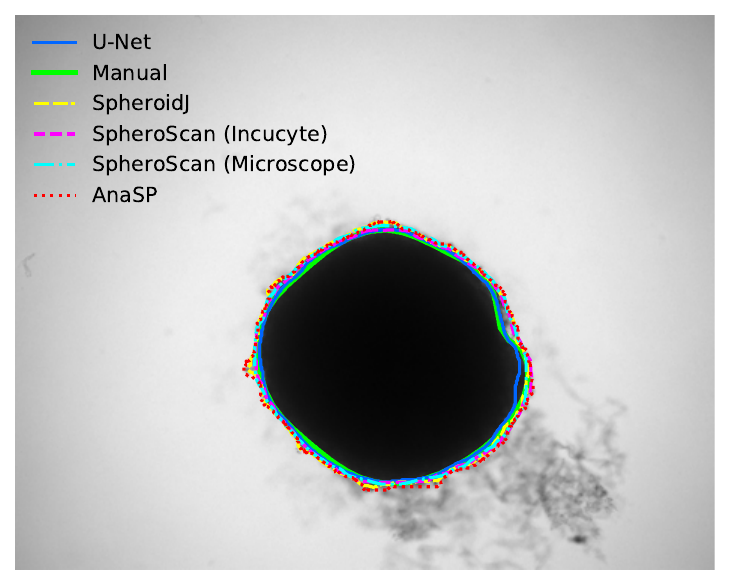}}
    \quad
    \subfloat[Treated spheroid with cell debris]{\includegraphics[width=0.49\linewidth]{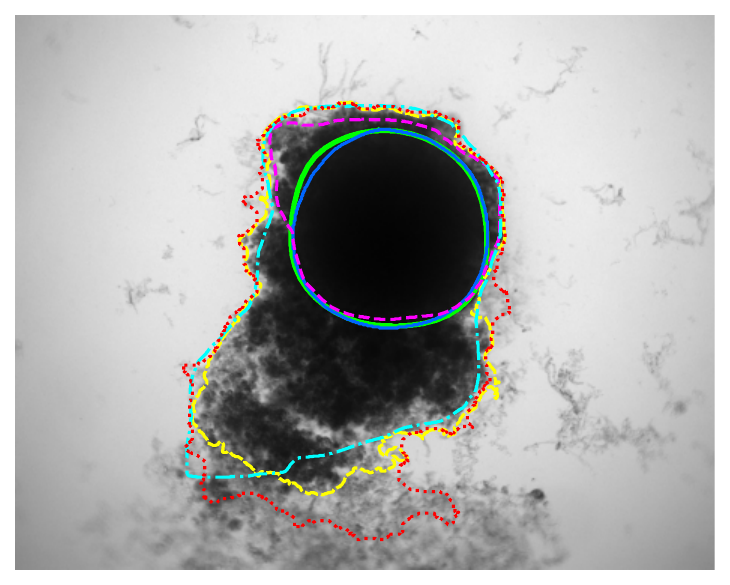}}
    \quad
    \caption{\label{sifig:model-comparison} Example images
      illustrating the current challenge of segmenting tumor
      spheroids: (a) While previously developed deep-learning models
      generalize excellently to new image data with
      well-distinguished, unobscured spheroids, typical for untreated
      cultures, (b) these models fail for cases of detached/relapsing
      spheroids after radiotherapy (one of the most common cancer
      treatments) due to debris and dead cells obscuring the spheroid.
      Shown are representative images of FaDu spheroids (a) without
      treatment and (b) several days after radiotherapy. Different
      segmentations are indicated by their outer contours. In
      particular, the manually set ground truth (green solid line) is
      compared to the four most recent deep-learning models provided
      by SpheroidJ~\cite{Lacetal2021} (HRNet-Seg with HRNet W30
      backbone, yellow close-dashed line),
      SpheroScan~\cite{AksKatAbeSheBesBurBigAdaMonGhe2023}
      (Region-based Convolutional Neural Network trained on images
      from IncuCyte Live-Cell Analysis System (purple dashed line) and
      ordinary microscope (torquoise dashed line) with recommended
      threshold 0.8), and
      AnaSP~\cite{PicPeiStePyuTumTazWevTesMarCas2023} (ResNet18, red
      dotted line). Only the U-Net presented in this manuscript (blue
      solid line) performs sufficiently in both scenarios. Size of
      each image corresponds to
      $2650 \ \mu\text{m} \times 2100 \ \mu\text{m}$ with a resolution
      of 1300 $\times$ 1030 pixel.}
\end{figure*}

\begin{table*}[!h]
  \centering
  \caption{\label{sitab:our-data} Statistical results reflecting the
    current challenge of segmenting tumor spheroids, corroborating
    \prettyref{sifig:model-comparison}: While previously developed
    deep-learning models generalize excellently to new image data with
    well-visible, unobscured spheroids (top line), these models fail
    for typical cases of detached/relapsing spheroids with debris of
    dead cells (bottom lines). Intersection over Union (IoU or Jaccard
    index) is reported (average (median) $\pm$ standard deviation) for
    each of the four most recent deep-learning models (provided by
    SpheroidJ~\cite{Lacetal2021},
    SpheroScan~\cite{AksKatAbeSheBesBurBigAdaMonGhe2023} (with
    recommended threshold 0.8), and
    AnaSP~\cite{PicPeiStePyuTumTazWevTesMarCas2023}) and each of our
    data sets. Only the U-Net presented in this manuscript performs
    sufficiently in both scenarios. Bold values highlight the optimum
    for each dataset.}
  \begin{tabular}{|lc|c||c|c|c|c|}
    \hline
    \multirow{2}{*}{Data set} & \multirow{2}{*}{\#{}Images} &
                                                             \multicolumn{5}{c|}{Intersection
                                                             over
                                                              Union
                                                              (IoU = 1-\gls{JCD})} \\
     &  & U-Net & SpheroidJ~\cite{Lacetal2021} &
                                                              SpheroScan
                                                              (Incu)~\cite{AksKatAbeSheBesBurBigAdaMonGhe2023}
    & SpheroScan (Micro)~\cite{AksKatAbeSheBesBurBigAdaMonGhe2023} & AnaSP~\cite{PicPeiStePyuTumTazWevTesMarCas2023} \\
    \hline
    \begin{tabular}{@{}r@{}} Spheroids without \\cell
      debris\end{tabular} & 200&$\mathbf{0.97 (0.97) \pm  0.01}$&$0.89 (0.90) \pm 0.05$&$0.90 (0.94)\pm 0.16$&$0.72 (0.92) \pm 0.38$&$0.81 (0.87) \pm 0.22$  \\
    \hline
    \hline
    % Training + Validation & 991 & $0.95 ( 0.96 ) \pm 0.05$ &$0.49 (0.49) \pm 0.27$ & $0.67 (0.86) \pm 0.35$ & $0.47 (0.50) \pm 0.37$ &$0.44 (0.40) \pm 0.28$ \\
    Training & 883 & $\mathbf{0.94 (0.95) \pm 0.05}$ &$0.51 (0.53) \pm 0.27$
        & $0.68 (0.86) \pm 0.34$ & $0.48 (0.59) \pm 0.37$ &$0.46(0.44) \pm 0.28$\\
    Validation & 108 & $\mathbf{0.95 (0.96) \pm 0.04}$ &$0.31 (0.29) \pm 0.19$ & $0.59 (0.80) \pm 0.39$ & $0.36(0.26) \pm 0.32$ &$0.28(0.25) \pm 0.19$ \\
    Test & 104 & $\mathbf{0.94 (0.96) \pm 0.06}$ & $0.41 (0.38) \pm
                                            0.27$&$0.66(0.88) \pm 0.36$&$0.48 (0.50) \pm 0.38$ &$0.35 (0.33) \pm 0.24$\\
    % Training w/o clean & 683 + 2 & $0.86 (0.92) \pm 0.16$ & $0.30 (0.14) \pm 0.29$ & $0.48 (0.58) \pm 0.42$ &$0.31 (0.06) \pm 0.38$ &$0.25 (0.10) \pm 0.27$ \\
    \hline
  \end{tabular}
\end{table*}

\begin{table*}
  \centering
  \caption{\label{sitab:their-data} Further validation of the trained
    U-Net on wide variety of published test data sets from previous
    deep-learning models. While the trained U-Net performs well on on
    roughly half of the data sets or 38\% of the images (with average
    IoU above 0.8), sometimes surpassing the original model
    corresponding to the data set, two types of images turn out
    problematic: (i) images with ambiguous ground truth (20\% of
    images) for which the U-Net may actually segment reasonably, see
    main text for detailed discussion, and (ii) images on which the
    spheroids appears semi-transparent (42\% of images), with
    individual cells being visible throughout the spheroid,
    potentially due to its small size or the chosen microscopy method.
    However, classical segmentation techniques work sufficiently well
    for both types of images (i) and (ii), making the use of
    deep-learning approaches in these cases unnecessary. For
    demonstration, the last two columns report the performance of the
    classical approach from the original publication and simple Otsu
    thresholding (sometimes after some image erosion for (ii))
    performed by us for images of types (i) and (ii). Note that the
    notation for data sets from
    Refs.~\cite{IvaParWalAleAshGelGar2014,Lacetal2021} stands for
    brightfield/fluorescence microscopy (B/F), Nikon Eclipse/Leica
    DMi8/Olympus microscope (N,L,O), 2x/5x/10x magnification (2/5/10),
    and suspension/collagen culture (S/C). Bold values highlight the
    optimum for each dataset.}
  \begin{tabular}{|ccc|c|c|cc|}
    \hline
    \multirow{3}{*}{Data set} & \multirow{3}{*}{\#{}Images} &
                                                             \multirow{3}{*}{Comments}
    & \multicolumn{4}{c|}{Intersection over Union (IoU = 1-\gls{JCD})} \\
    \cline{4-7}
    & & & \multicolumn{2}{c|}{deep-learning model} &
                                                     \multicolumn{2}{c|}{classical segmentation} \\
             &  &  & Original & U-Net & Original & Otsu thresholding \\
    \hline
        BO10S~\cite{IvaParWalAleAshGelGar2014} & 66 & & $0.92 \pm
                                                    0.03$~\cite{Lacetal2021}&$\mathbf{0.97(0.98)\pm0.05}$
    &  $0.94 \pm 0.03$~\cite{Lacetal2021} & \\
    \hline
    % BN10S & 105 & 3x, good on first 21, boundary issue (Otsu: $0.84 (0.85) \pm 0.13$)&$0.97 \pm 0.01$&$0.67 (0.66) \pm 0.18$ \\
    \multirow{2}{*}{BN10S~\cite{Lacetal2021}} & 21 & & \multirow{2}{*}{$\mathbf{0.97 \pm 0.01}$}
    &$0.89 (0.89) \pm 0.01$ & \multirow{2}{*}{$0.95 \pm 0.01$}& \multirow{2}{*}{$0.84 (0.85) \pm 0.13$} \\
                             & 84 & ambiguous ground truth & &$0.61 (0.62) \pm 0.15$ & & \\
    \hline
    BN2S~\cite{Lacetal2021} & 154 & semi-transparent & $\mathbf{0.96 \pm 0.01}$ & $0.30 (0.27)\pm 0.22$ &
                                                                     $0.94
                                                                     \pm
                                                                     0.02$
        & $0.87 (0.95) \pm 0.27$   \\ %small - thin -bubbly
    \hline
        BL5S~\cite{Lacetal2021} & 50 & semi-transparent & $0.75 \pm 0.25$
    & $0.37 (0.13) \pm 0.39$ & $0.64 \pm 0.30$ & $\mathbf{0.85 (0.93) \pm
                                                 0.20}$ \\
    \hline
    % FN2S & 34 & one image with double spheroid, 4 small spheroids with
    %             bubbly structure &$0.78 \pm 0.20$ & $0.81 (0.93) \pm
    %                                                 0.30$ \\
    \multirow{2}{*}{FN2S~\cite{Lacetal2021}} & 30 & &
                                                      \multirow{2}{*}{$0.78
                                                      \pm 0.20$} &
                                                                   $0.91
                                                                   (0.94)
                                                                   \pm
                                                                   0.08$
      & \multirow{2}{*}{$\mathbf{0.82 \pm 0.17}$} &  \\
     & 4 & semi-transparent & & $0$ & & $0.83(0.83) \pm 0.02$ \\
\hline
    % FL5C & 19 & good for 1,3,4,5, boundary issue (Otsu: $0.73 (0.80) \pm 0.21$)  & $0.71 \pm 0.30$ &$0.55 (0.51) \pm 0.24$\\
    \multirow{2}{*}{FL5C\cite{Lacetal2021}} & 4 & &
                                                    \multirow{2}{*}{$0.71
                                                    \pm 0.30$} &$0.92
                                                                 (0.91)
                                                                 \pm
                                                                 0.01$
      & \multirow{2}{*}{$0.67 \pm 0.17$} & \\
     & 15 & ambiguous ground truth & &$0.45 (0.45) \pm 0.16$ & &
                                                                       $\mathbf{0.80
                                                                   (0.85)
                                                                   \pm
                                                                   0.11}$\\
\hline
    FL5S~\cite{Lacetal2021} & 50 & & $0.70 \pm 0.26$ & $\mathbf{0.91(0.91) \pm
                                                       0.05}$ &
                                                               $0.89 \pm
                                                               0.07$ & \\
\hline
    \multirow{2}{*}{AnaSP~\cite{PicPeiStePyuTumTazWevTesMarCas2023}} &
                                                                       16
                                                           &  &
                                                                \multirow{2}{*}{$\sim 0.92$} &
                                                                  $0.97 (0.98) \pm 0.02$ & & \\
     & 2 & semi-transparent & & $0.01 (0.01) \pm 0.01$ & & $0.97(0.97) \pm
                                                           0.00$ \\
    \hline
  \end{tabular}
\end{table*}

\begin{figure*}
  \centering
  \includegraphics[width=0.28\linewidth,trim=2cm 0cm 2cm 0cm,clip]{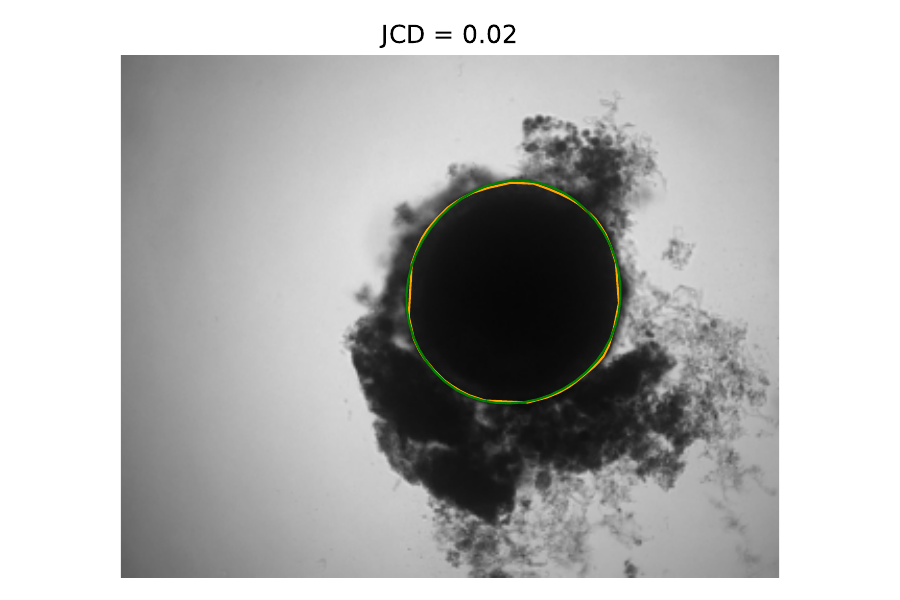}
  \includegraphics[width=0.28\linewidth,trim=2cm 0cm 2cm 0cm,clip]{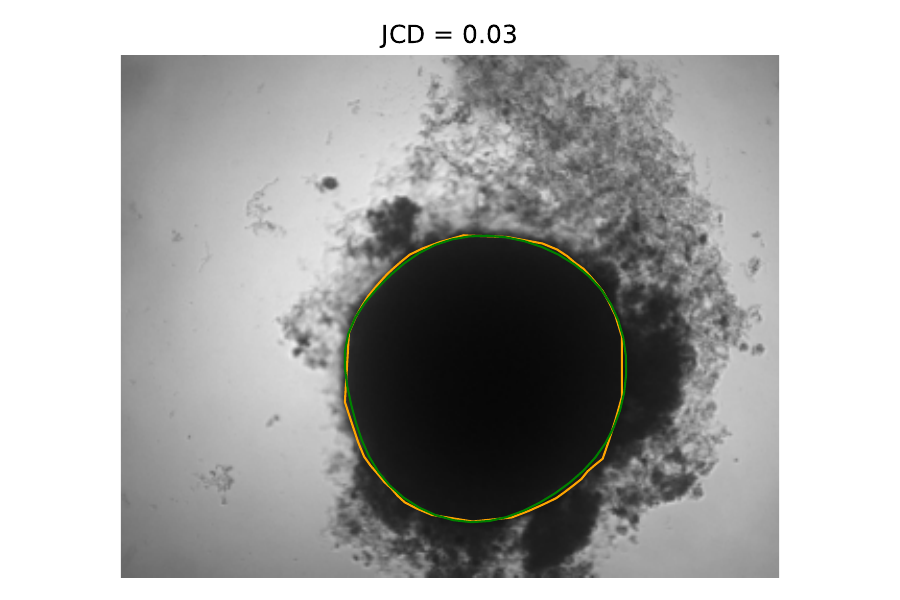}
    \includegraphics[width=0.28\linewidth,trim=2cm 0cm 2cm 0cm,clip]{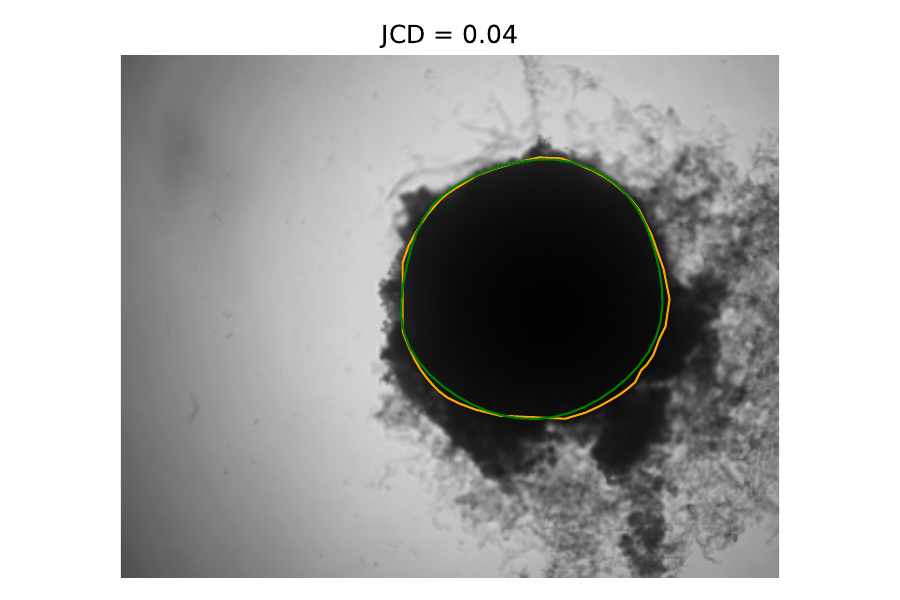}

  \includegraphics[width=0.28\linewidth,trim=2cm 0cm 2cm 0cm,clip]{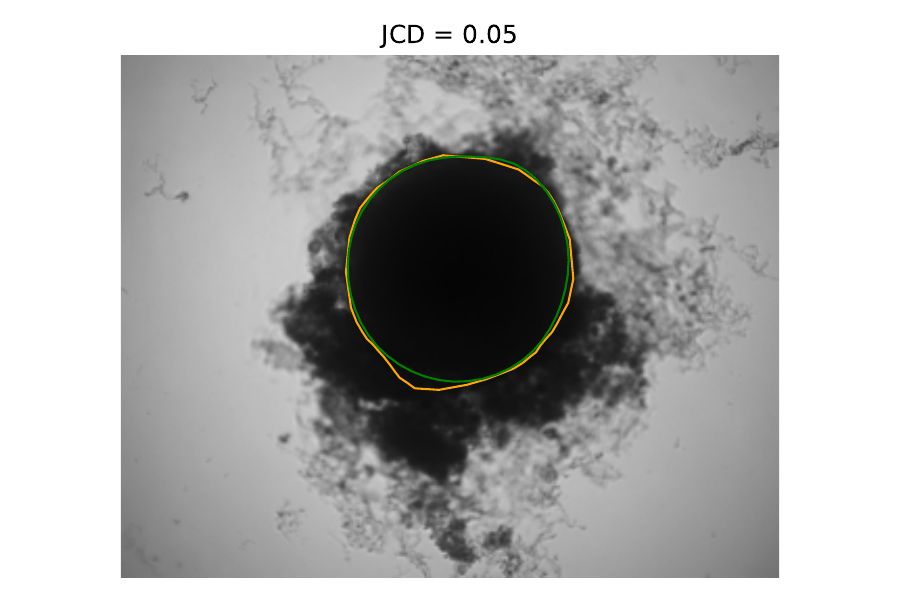}
  \includegraphics[width=0.28\linewidth,trim=2cm 0cm 2cm 0cm,clip]{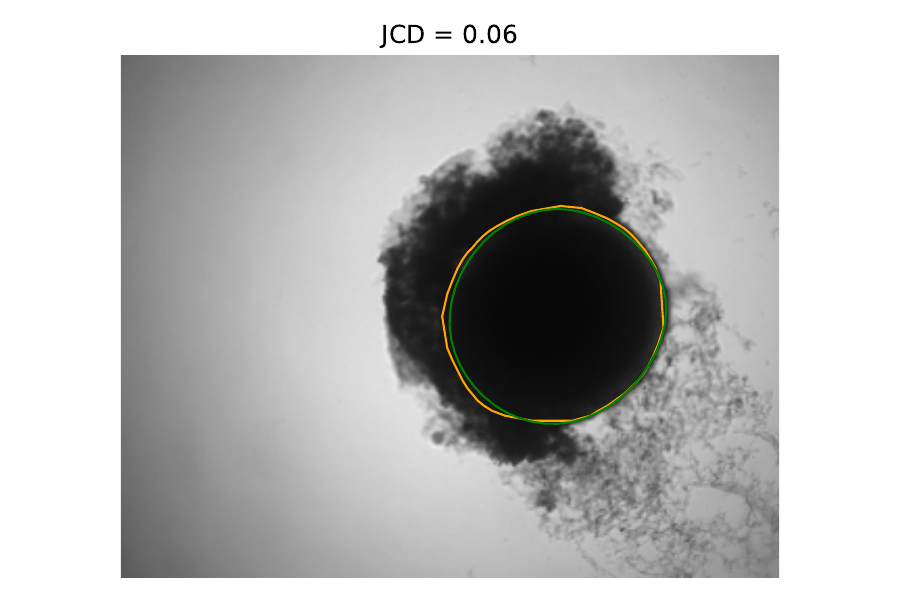}
      \includegraphics[width=0.28\linewidth,trim=2cm 0cm 2cm 0cm,clip]{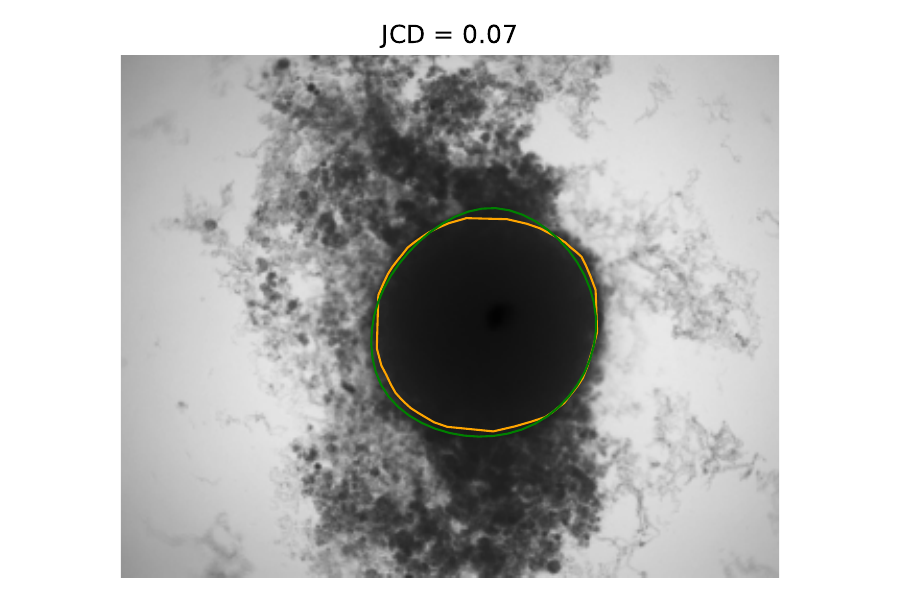}

  \includegraphics[width=0.28\linewidth,trim=2cm 0cm 2cm 0cm,clip]{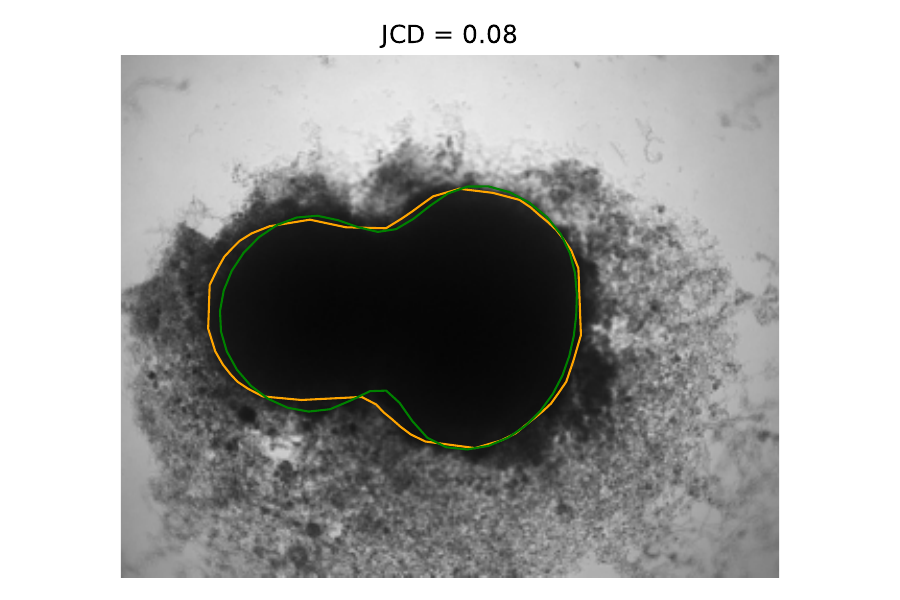}
  \includegraphics[width=0.28\linewidth,trim=2cm 0cm 2cm 0cm,clip]{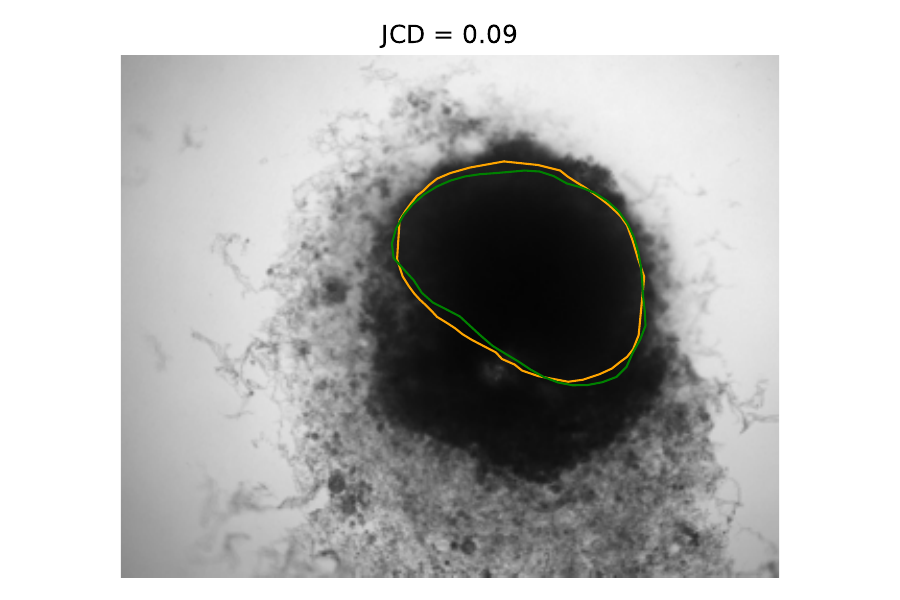}
      \includegraphics[width=0.28\linewidth,trim=2cm 0cm 2cm 0cm,clip]{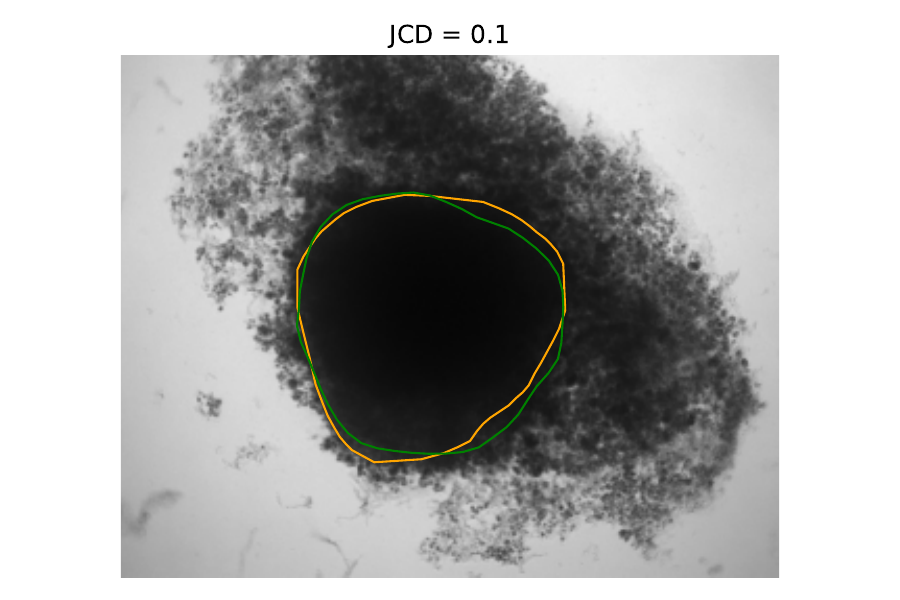}
      \caption{Selection of representative images as optical reference
        for \gls{JCD}$\leq 0.1$ with automatic segmentation from the
        optical U-Net (orange) and manual segmentation from biological
        expert H2 (green). From the tested images, $52 \%$ fall into
        this range of \gls{JCD}$\leq 0.1$ and $67 \%$ when only
        spheroids beyond the standard size $d_T>400 \ \mu$m are
        considered. Size of each image corresponds to
        $2650 \ \mu\text{m} \times 2100 \ \mu\text{m}$. Note that the
        U-Net is trained on an independent manual segmentations from
        another biological expert H1. Images are selected from the
        extended validation data set (\prettyref{fig:validation}), in
        particular, from the subset of larger, better discriminated
        spheroids at intermediate levels of debris for illustration.
        Note that in the example for $\gls{JCD}=0.08$, the two
        attached spheroids are correctly segmented, although the
        training data set does not contain such cases.}
    \label{sifig:samplesJCD10}
\end{figure*}

\begin{figure*}
  \centering
  \includegraphics[width=0.28\linewidth,trim=2cm 0cm 2cm 0cm,clip]{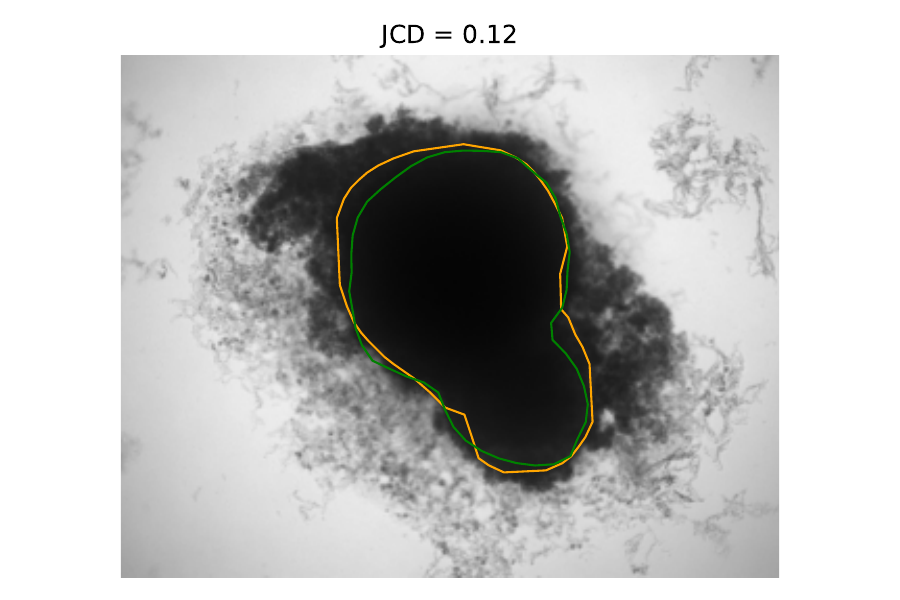}
  \includegraphics[width=0.28\linewidth,trim=2cm 0cm 2cm 0cm,clip]{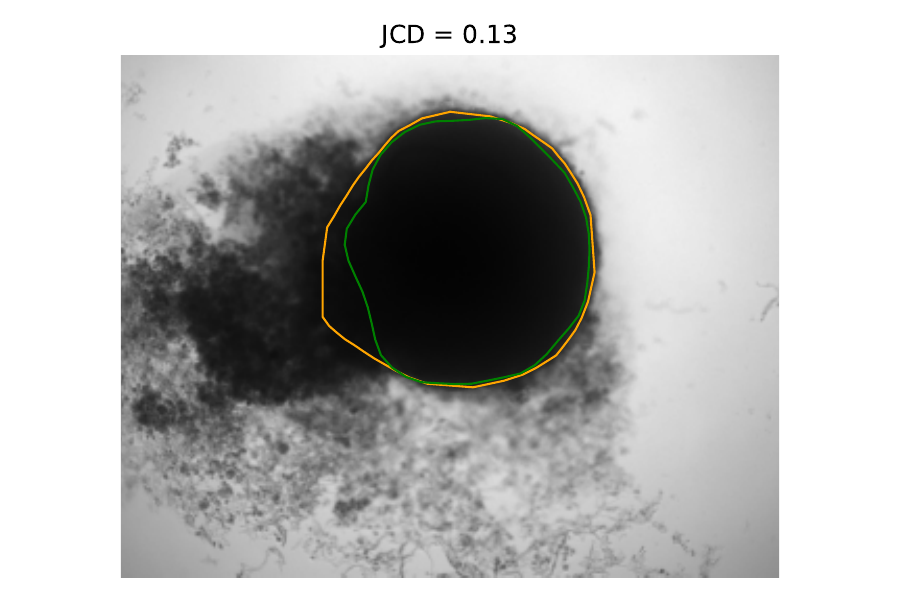}
    \includegraphics[width=0.28\linewidth,trim=2cm 0cm 2cm 0cm,clip]{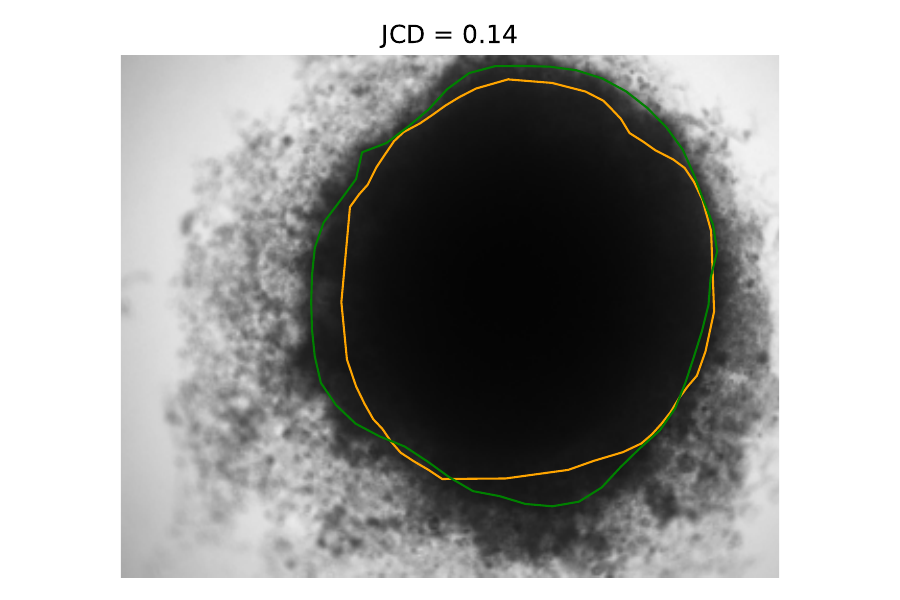}

  \includegraphics[width=0.28\linewidth,trim=2cm 0cm 2cm 0cm,clip]{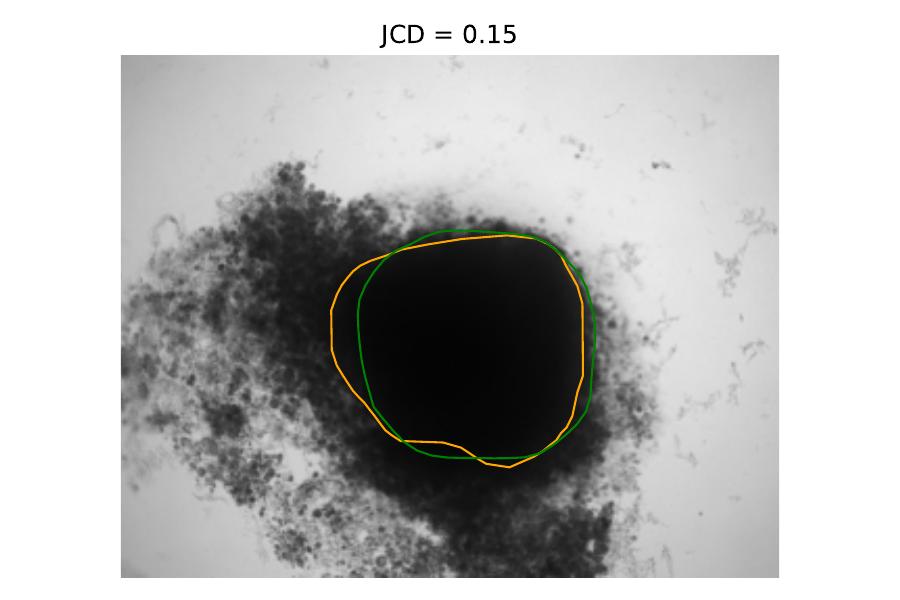}
  \includegraphics[width=0.28\linewidth,trim=2cm 0cm 2cm 0cm,clip]{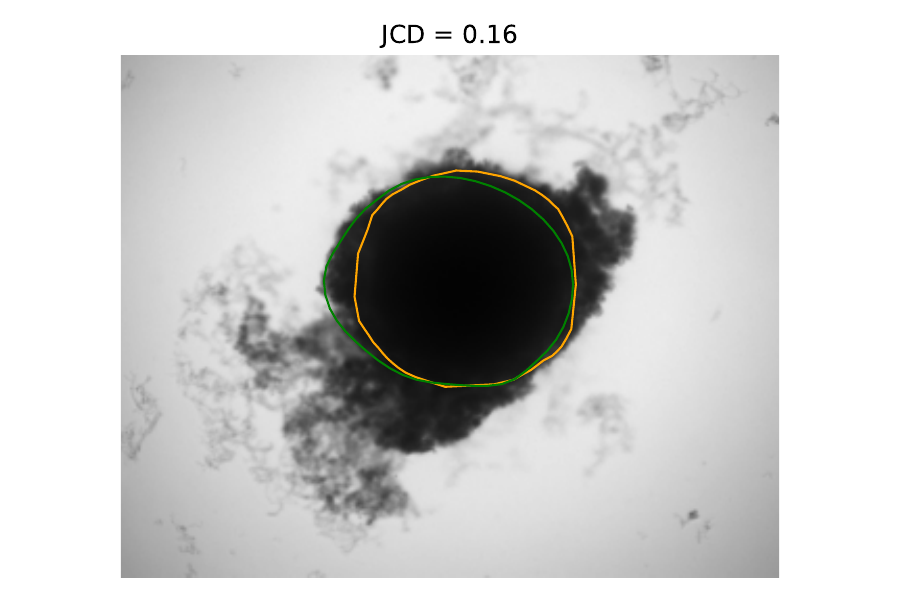}
      \includegraphics[width=0.28\linewidth,trim=2cm 0cm 2cm 0cm,clip]{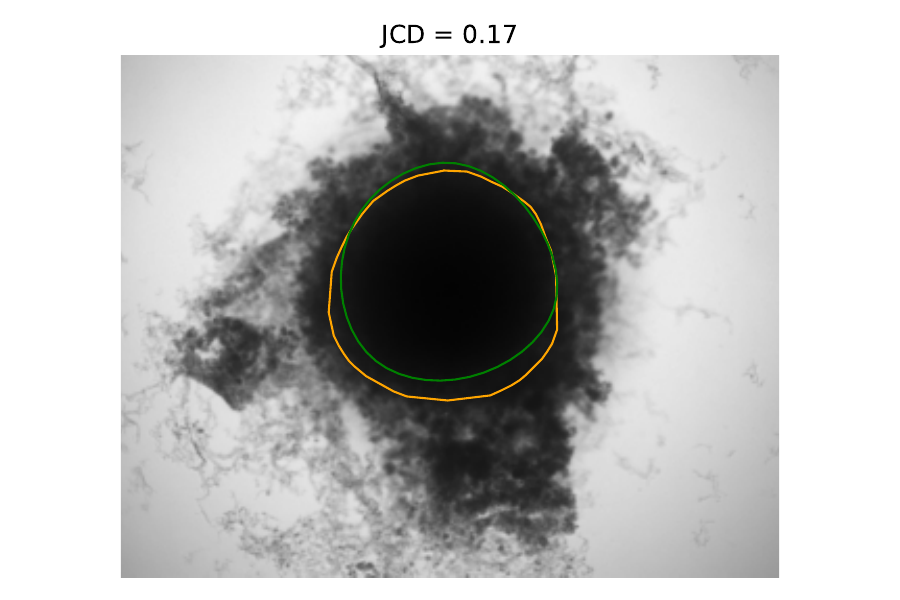}

  \includegraphics[width=0.28\linewidth,trim=2cm 0cm 2cm 0cm,clip]{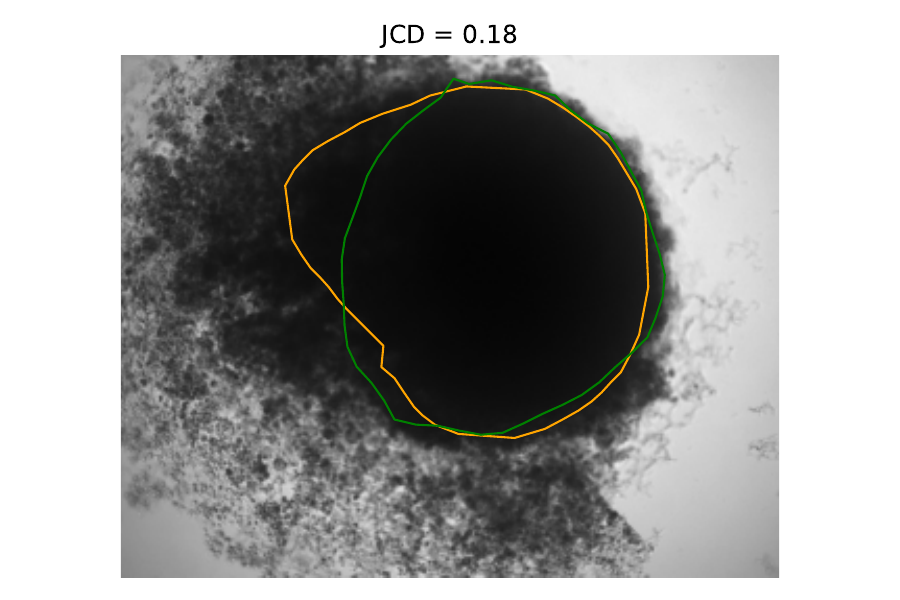}
  \includegraphics[width=0.28\linewidth,trim=2cm 0cm 2cm 0cm,clip]{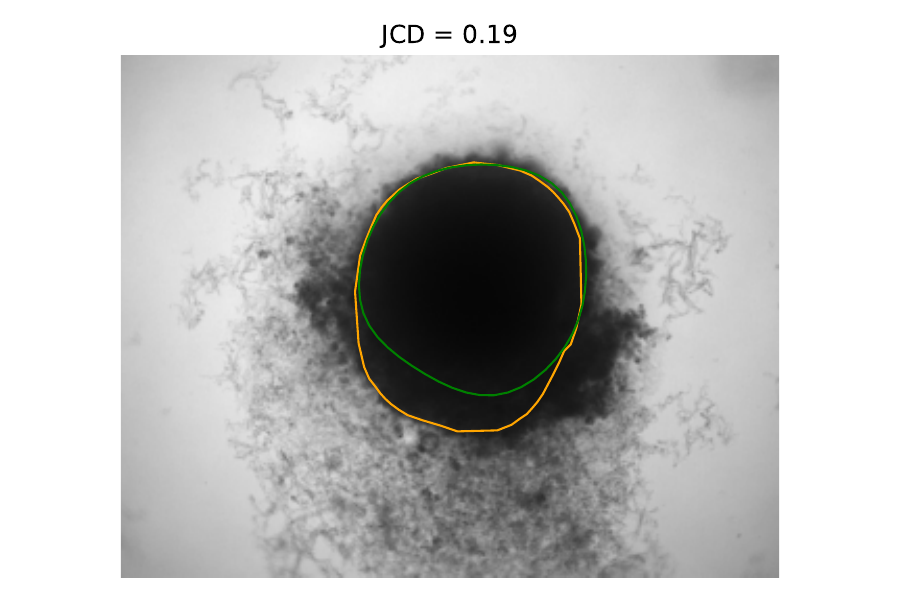}
      \includegraphics[width=0.28\linewidth,trim=2cm 0cm 2cm 0cm,clip]{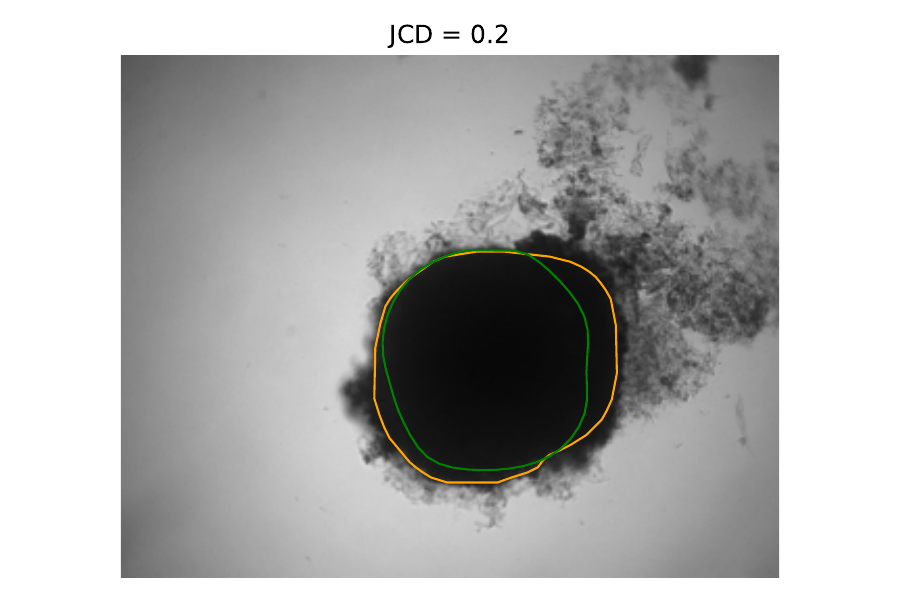}
      \caption{Selection of representative images as optical reference
        for $0.1<\gls{JCD}\leq 0.2$ with automatic segmentation from
        the optical U-Net (orange) and manual segmentation from
        biological expert H2 (green), analogous to
        \prettyref{sifig:samplesJCD10}. From the tested images,
        $22 \%$ fall into the range $0.1<\gls{JCD}\leq 0.2$ and
        $23 \%$ when only spheroids beyond the standard size
        $d_T>400 \ \mu$m are considered. Note that in the example of
        $\gls{JCD}=0.12$, the two attached spheroids are correctly
        segmented, although the training data set does not contain
        such cases.}
    \label{sifig:samplesJCD20}
\end{figure*}

\begin{figure*}
  \centering
  \includegraphics[width=0.28\linewidth,trim=2cm 0cm 2cm 0cm,clip]{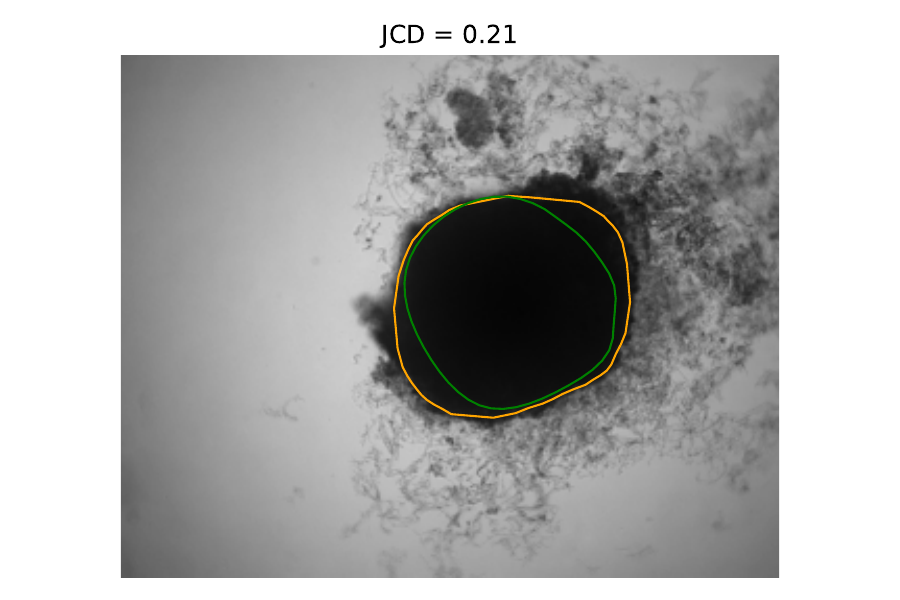}
  \includegraphics[width=0.28\linewidth,trim=2cm 0cm 2cm 0cm,clip]{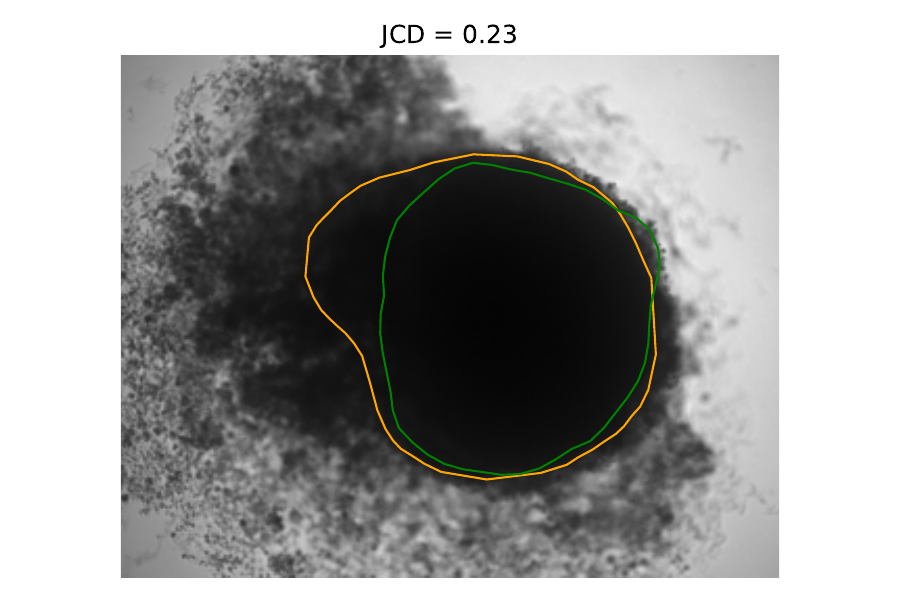}
    \includegraphics[width=0.28\linewidth,trim=2cm 0cm 2cm 0cm,clip]{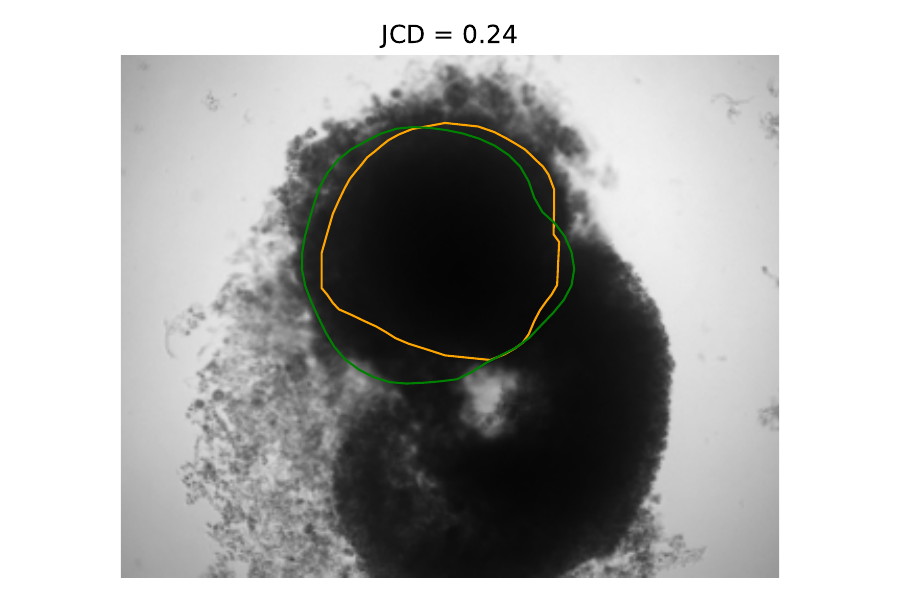}

  \includegraphics[width=0.28\linewidth,trim=2cm 0cm 2cm 0cm,clip]{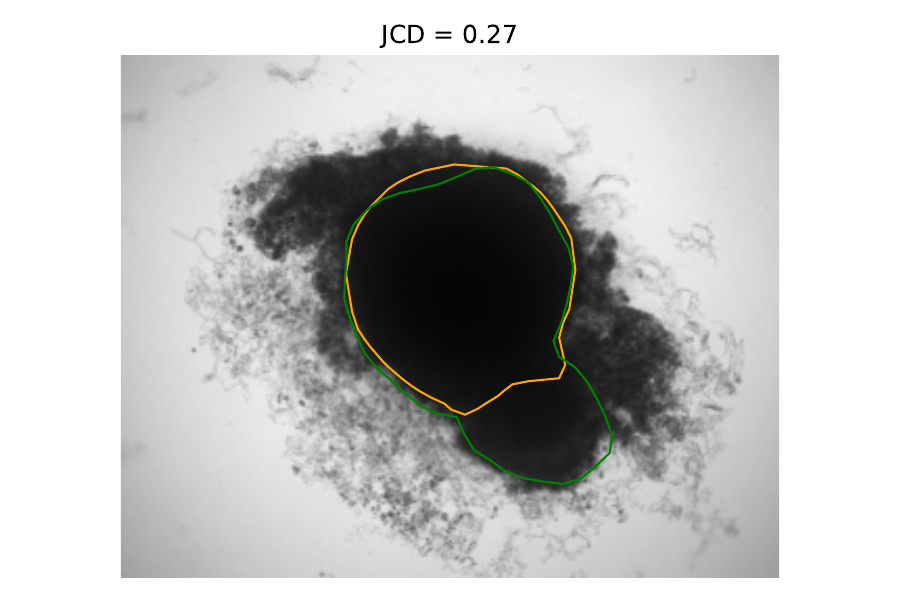}
  \includegraphics[width=0.28\linewidth,trim=2cm 0cm 2cm 0cm,clip]{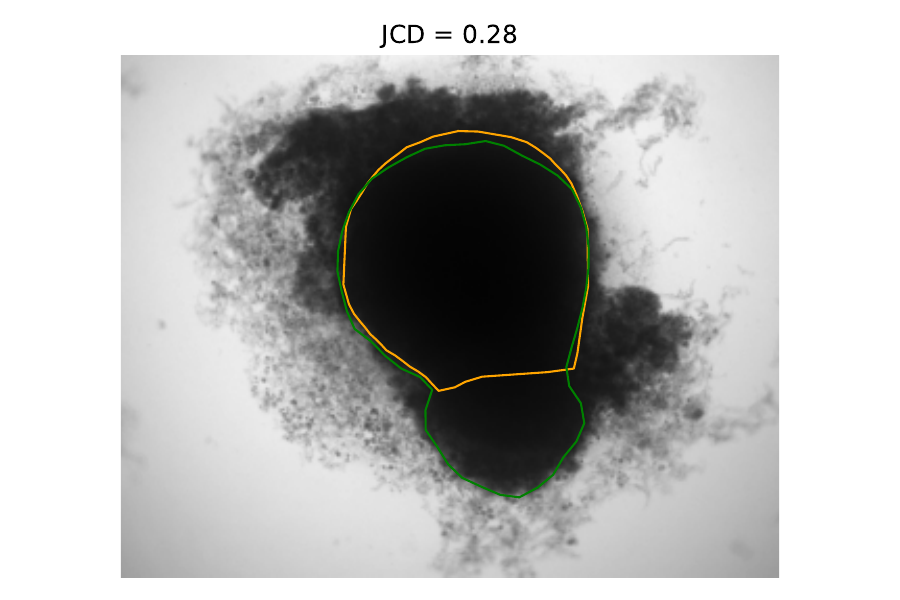}
      \includegraphics[width=0.28\linewidth,trim=2cm 0cm 2cm 0cm,clip]{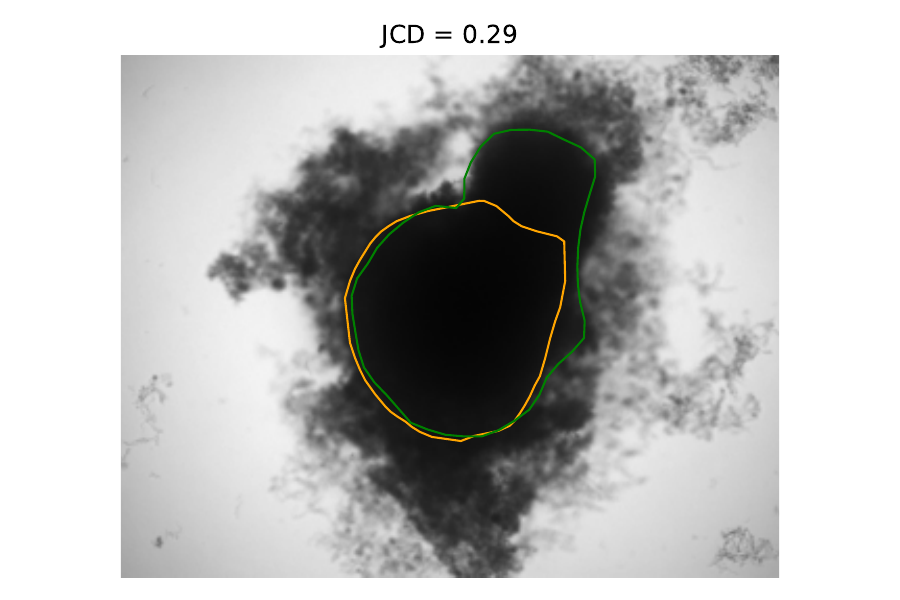}

  \includegraphics[width=0.28\linewidth,trim=2cm 0cm 2cm 0cm,clip]{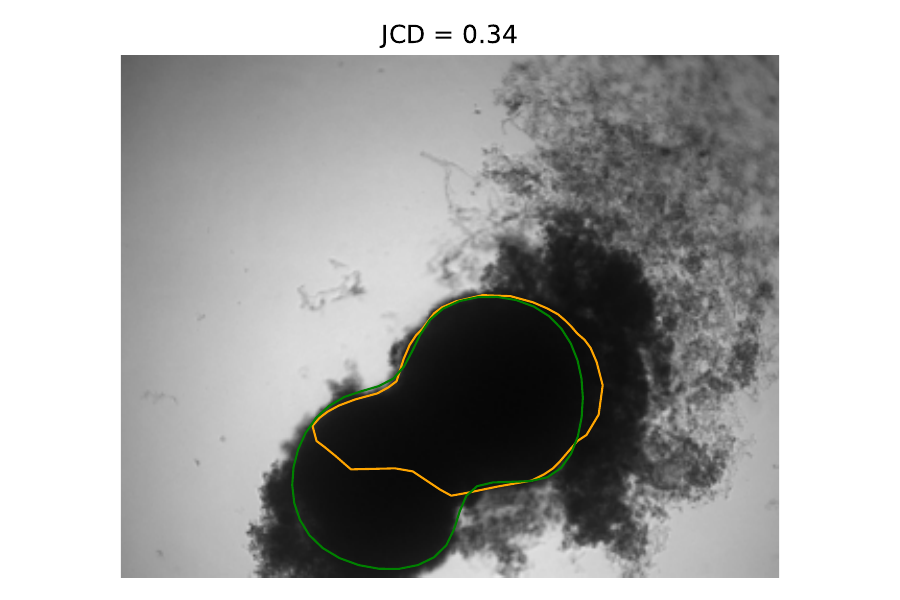}
  \includegraphics[width=0.28\linewidth,trim=2cm 0cm 2cm 0cm,clip]{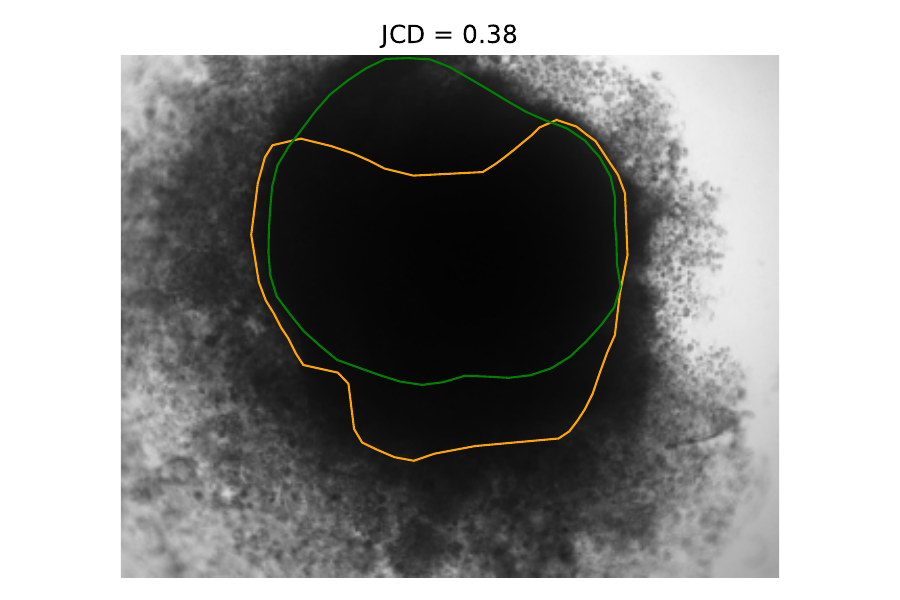}
      \includegraphics[width=0.28\linewidth,trim=2cm 0cm 2cm 0cm,clip]{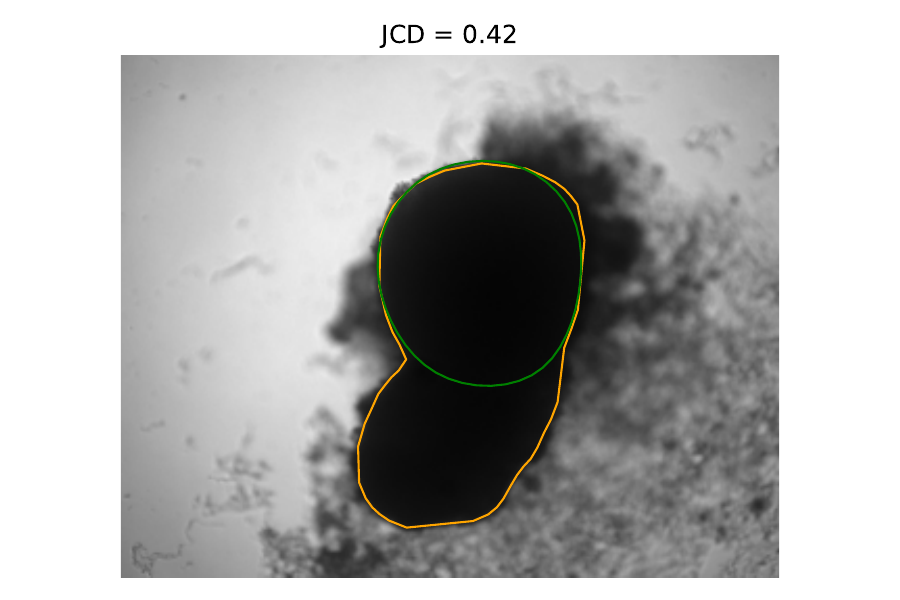}
      \caption{Selection of representative images as optical reference
        for larger deviations $\gls{JCD}> 0.2$ of the automatic
        segmentation from the optical U-Net (orange) and manual
        segmentation from biological expert H2 (green), analogous to
        \prettyref{sifig:samplesJCD10}. From the tested images,
        $26 \%$ fall into the range $\gls{JCD}> 0.2$, but merely
        $10 \%$ when only spheroids beyond the standard size
        $d_T>400 \ \mu$m are considered. Note that deviations at
        larger spheroids are often due to cases of double-spheroids,
        e.g., $\gls{JCD}=0.27, 0.28, 0.29, 0.34, 0.42$, which are
        inconsistently recognized as either one or two spheroids even
        by the human.}
    \label{sifig:samplesJCD30}
\end{figure*}

\begin{figure}[!ht]
    \centering
    \subfloat{\includegraphics[width=\linewidth]{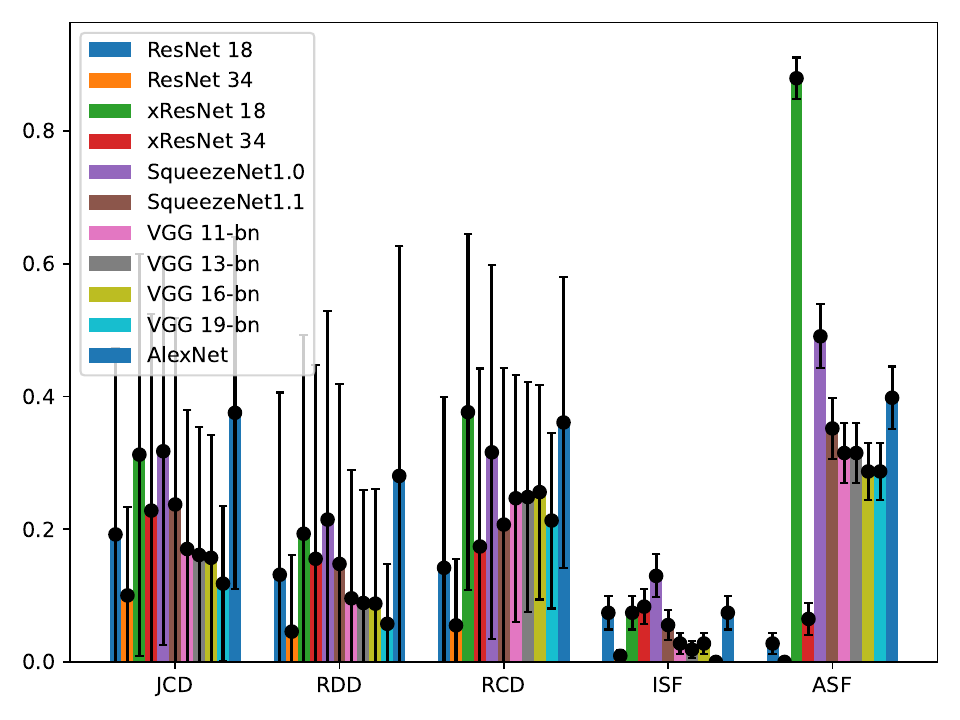}}

    \subfloat{
        \begin{tabular}{|l|r|r|r|r|r|}
            \hline
            Backbone & \gls{JCD} & \gls{DD} & \gls{CD} & \gls{ISF} & \gls{ASF} \\
            \hline
            \hline
            ResNet 18 & 0.192 & 0.131 & 0.142 & 0.074 & 0.028 \\
            $\pm$ & 0.282 & 0.275 & 0.258 & 0.025 & 0.016 \\
            \hline
            ResNet 34 & \cellcolor{gray!25}\textbf{0.100} & \cellcolor{gray!25}\textbf{0.046} & \cellcolor{gray!25}\textbf{0.055} & 0.009 & \cellcolor{gray!25}\textbf{0.000} \\
            $\pm$ & 0.133 & 0.116 & 0.100 & 0.009 & 0.000 \\
            \hline
            xResNet 18 & 0.312 & 0.193 & 0.376 & 0.074 & 0.880 \\
            $\pm$ & 0.303 & 0.300 & 0.268 & 0.025 & 0.031 \\
            \hline
            xResNet 34 & 0.228 & 0.155 & 0.174 & 0.083 & 0.065 \\
            $\pm$ & 0.297 & 0.292 & 0.268 & 0.027 & 0.024 \\
            \hline
            SqueezeNet1.0 & 0.317 & 0.215 & 0.316 & 0.130 & 0.491 \\
            $\pm$ & 0.292 & 0.315 & 0.282 & 0.032 & 0.048 \\
            \hline
            SqueezeNet1.1 & 0.237 & 0.148 & 0.207 & 0.056 & 0.352 \\
            $\pm$ & 0.281 & 0.271 & 0.237 & 0.022 & 0.046 \\
            \hline
            VGG 11-bn & 0.170 & 0.096 & 0.247 & 0.028 & 0.315 \\
            $\pm$ & 0.210 & 0.193 & 0.186 & 0.016 & 0.045 \\
            \hline
            VGG 13-bn & 0.161 & 0.089 & 0.248 & 0.019 & 0.315 \\
            $\pm$ & 0.193 & 0.170 & 0.173 & 0.013 & 0.045 \\
            \hline
            VGG 16-bn & 0.157 & 0.088 & 0.256 & 0.028 & 0.287 \\
            $\pm$ & 0.185 & 0.173 & 0.162 & 0.016 & 0.044 \\
            \hline
            VGG 19-bn & 0.118 & 0.057 & 0.213 & \cellcolor{gray!25}\textbf{0.000} & 0.287 \\
            $\pm$ & 0.117 & 0.090 & 0.133 & 0.000 & 0.044 \\
            \hline
            AlexNet & 0.375 & 0.280 & 0.361 & 0.074 & 0.398 \\
            $\pm$ & 0.265 & 0.346 & 0.219 & 0.025 & 0.047 \\
            \hline
        \end{tabular}
    }
    \caption{Values of evaluation metrics for different backbones of
      the U-Net. Bold values highlight the optimum in each column. The
      other hyperparameters are fixed at Optimizer: Adam, Loss:
      Cross-Entropy, Resize factor: 1/2, Transfer learning: Yes, Data
      augmentation: No. Overall the ResNet 34 achieved the best
      results and is picked as the backbone for the U-Net.}
    \label{sifig:backbone_unet}
\end{figure}

\begin{figure}[!ht]
    \centering
    \subfloat{\includegraphics[width=\linewidth]{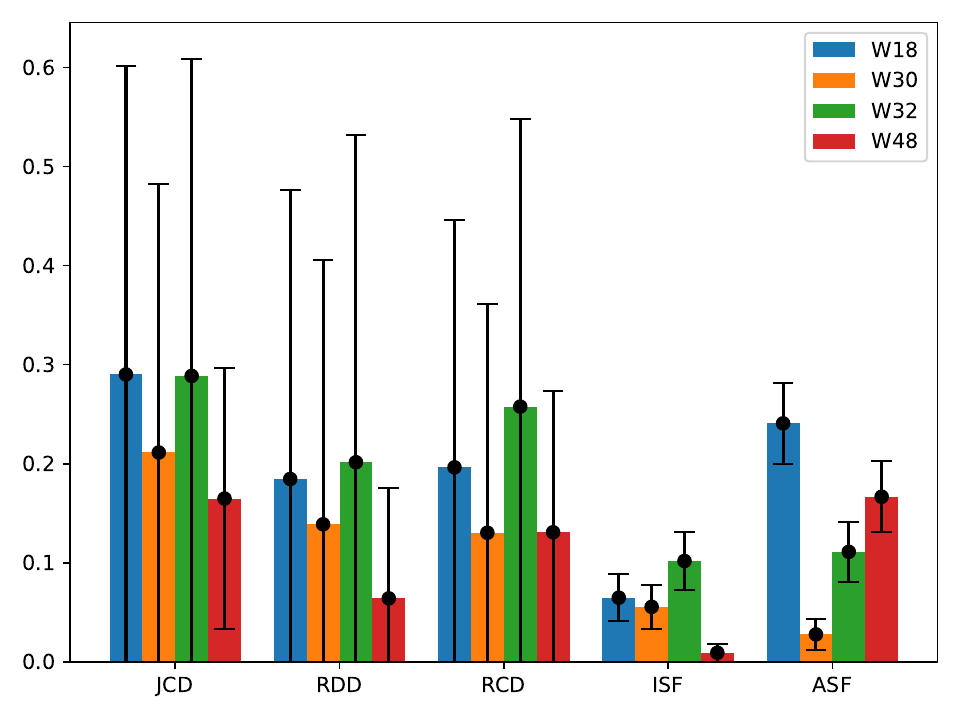}}

    \subfloat{
        \begin{tabular}{|l|r|r|r|r|r|}
            \hline
            Backbone & \gls{JCD} & \gls{DD} & \gls{CD} & \gls{ISF} & \gls{ASF} \\
            \hline
            \hline
            W18 & 0.290 & 0.185 & 0.196 & 0.065 & 0.241 \\
            $\pm$ & 0.311 & 0.291 & 0.249 & 0.024 & 0.041 \\
            \hline
            W30 & 0.211 & 0.139 & \cellcolor{gray!25}\textbf{0.130} & 0.056 & \cellcolor{gray!25}\textbf{0.028} \\
            $\pm$ & 0.271 & 0.266 & 0.231 & 0.022 & 0.016 \\
            \hline
            W32 & 0.289 & 0.201 & 0.258 & 0.102 & 0.111 \\
            $\pm$ & 0.320 & 0.330 & 0.290 & 0.029 & 0.030 \\
            \hline
            W48 & \cellcolor{gray!25}\textbf{0.165} & \cellcolor{gray!25}\textbf{0.064} & 0.131 & \cellcolor{gray!25}\textbf{0.009} & 0.167 \\
            $\pm$ & 0.132 & 0.111 & 0.143 & 0.009 & 0.036 \\
            \hline
        \end{tabular}
    }
    \caption{Values of evaluation metrics for different backbones of
      the HRNet. Bold values highlight the optimum in each column. The
      other hyperparameters are fixed at Optimizer: Adam, Loss:
      Cross-Entropy, Resize factor: 1/2, Transfer learning: Yes, Data
      augmentation: No. Overall the W48 achieved the best results and
      is picked as the backbone for the HRNet.}
    \label{sifig:backbone_hrnet}
\end{figure}

\begin{figure}[!ht]
    \centering
    \subfloat{\includegraphics[width=\linewidth]{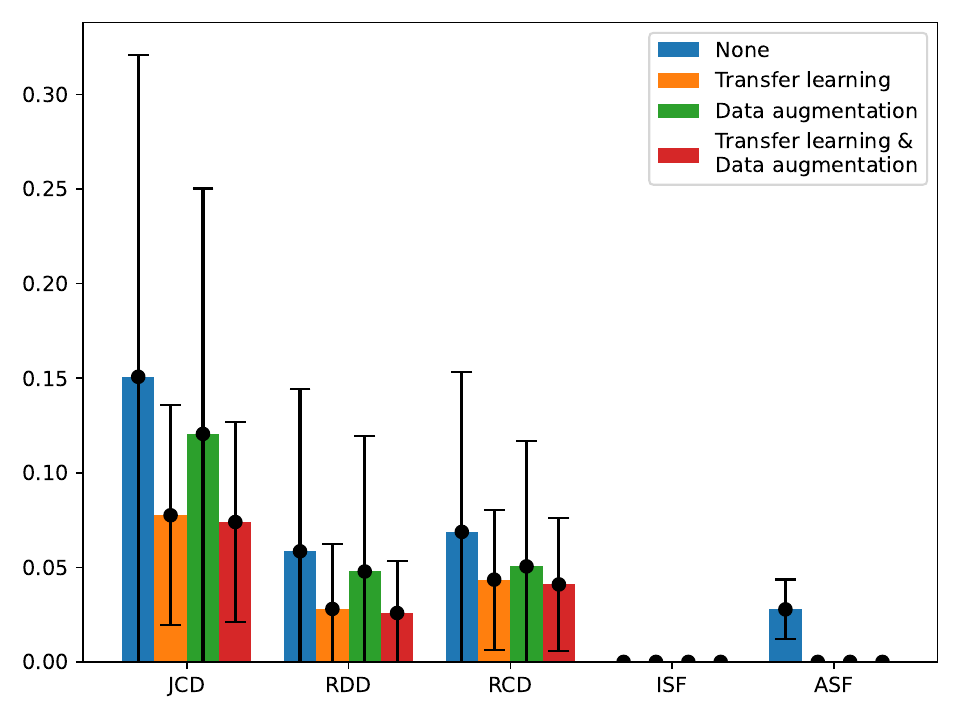}}

    \subfloat{
        \begin{tabular}{|l|r|r|r|r|r|}
            \hline
            Extension & \gls{JCD} & \gls{DD} & \gls{CD} & \gls{ISF} & \gls{ASF} \\
            \hline
            \hline
            None & 0.151 & 0.058 & 0.069 & \cellcolor{gray!25}\textbf{0.000} & 0.028 \\
            $\pm$ & 0.170 & 0.086 & 0.084 & 0.000 & 0.016 \\
            \hline
            Transfer learning & 0.078 & 0.028 & 0.043 & \cellcolor{gray!25}\textbf{0.000} & \cellcolor{gray!25}\textbf{0.000} \\
            $\pm$ & 0.058 & 0.034 & 0.037 & 0.000 & 0.000 \\
            \hline
            Data augmentation & 0.121 & 0.048 & 0.050 & \cellcolor{gray!25}\textbf{0.000} & \cellcolor{gray!25}\textbf{0.000} \\
            $\pm$ & 0.130 & 0.072 & 0.066 & 0.000 & 0.000 \\
            \hline
            Transfer learning \& & \cellcolor{gray!25}\textbf{0.074} & \cellcolor{gray!25}\textbf{0.026} & \cellcolor{gray!25}\textbf{0.041} & \cellcolor{gray!25}\textbf{0.000} & \cellcolor{gray!25}\textbf{0.000} \\
            Data augmentation $\pm$ & 0.053 & 0.027 & 0.035 & 0.000 & 0.000 \\
            \hline
        \end{tabular}
    }
    \caption{Values of evaluation metrics for different extensions of
      the training data set, which is used by the U-Net. Bold values
      highlight the optimum in each column. The other hyperparameters
      are fixed at Backbone: ResNet 34, Optimizer: Adam, Loss:
      Cross-Entropy, Resize factor: 1/2. The accuracy of the U-Net is
      highest when transfer learning and data augmentation is used.}
    \label{sifig:extension_unet}
\end{figure}

\begin{figure}[!ht]
    \centering
    \subfloat{\includegraphics[width=\linewidth]{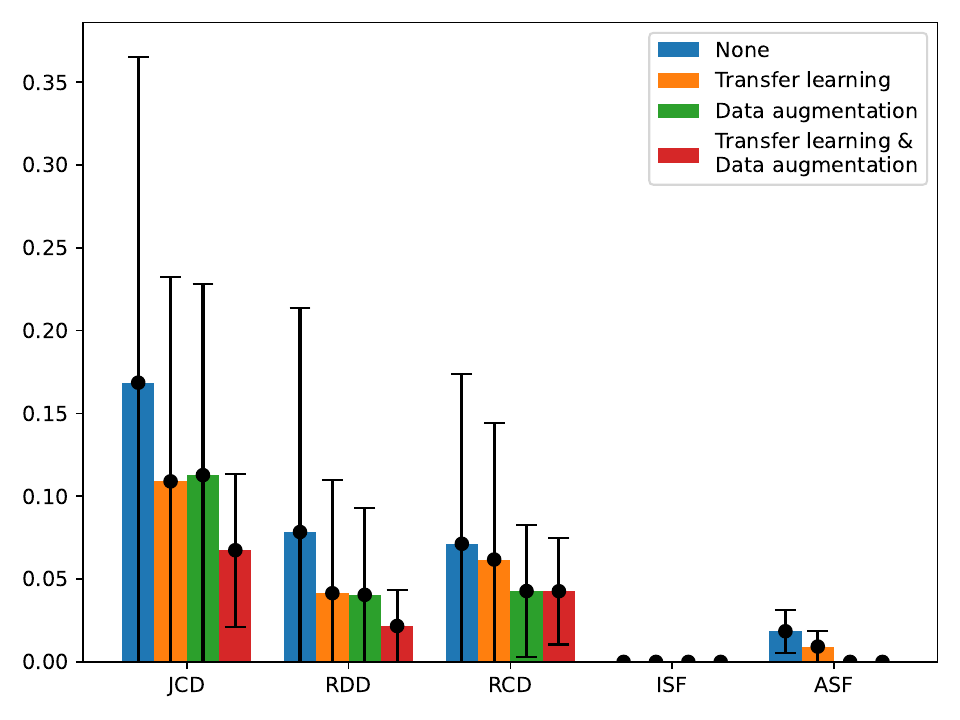}}

    \subfloat{
        \begin{tabular}{|l|r|r|r|r|r|}
            \hline
            Extension & \gls{JCD} & \gls{DD} & \gls{CD} & \gls{ISF} & \gls{ASF} \\
            \hline
            \hline
            None & 0.169 & 0.078 & 0.071 & \cellcolor{gray!25}\textbf{0.000} & 0.019 \\
            $\pm$ & 0.196 & 0.135 & 0.103 & 0.000 & 0.013 \\
            \hline
            Transfer learning & 0.109 & 0.041 & 0.062 & \cellcolor{gray!25}\textbf{0.000} & 0.009 \\
            $\pm$ & 0.123 & 0.069 & 0.082 & 0.000 & 0.009 \\
            \hline
            Data augmentation & 0.113 & 0.040 & \cellcolor{gray!25}\textbf{0.043} & \cellcolor{gray!25}\textbf{0.000} & \cellcolor{gray!25}\textbf{0.000} \\
            $\pm$ & 0.115 & 0.052 & 0.040 & 0.000 & 0.000 \\
            \hline
            Transfer learning \& & \cellcolor{gray!25}\textbf{0.067} & \cellcolor{gray!25}\textbf{0.022} & \cellcolor{gray!25}\textbf{0.043} & \cellcolor{gray!25}\textbf{0.000} & \cellcolor{gray!25}\textbf{0.000} \\
            Data augmentation $\pm$ & 0.046 & 0.022 & 0.032 & 0.000 & 0.000 \\
            \hline
        \end{tabular}
    }
    \caption{Values of evaluation metrics for different extensions of
      the training data set, which is used by the HRNet. Bold values
      highlight the optimum in each column. The other hyperparameters
      are fixed at Backbone: W48, Optimizer: Adam, Loss:
      Cross-Entropy, Resize factor: 1/2. The accuracy of the HRNet is
      highest when transfer learning and data augmentation is used.}
    \label{sifig:extension_hrnet}
  \end{figure}

  \begin{figure}[!ht]
    \centering
    \subfloat{\includegraphics[width=\linewidth]{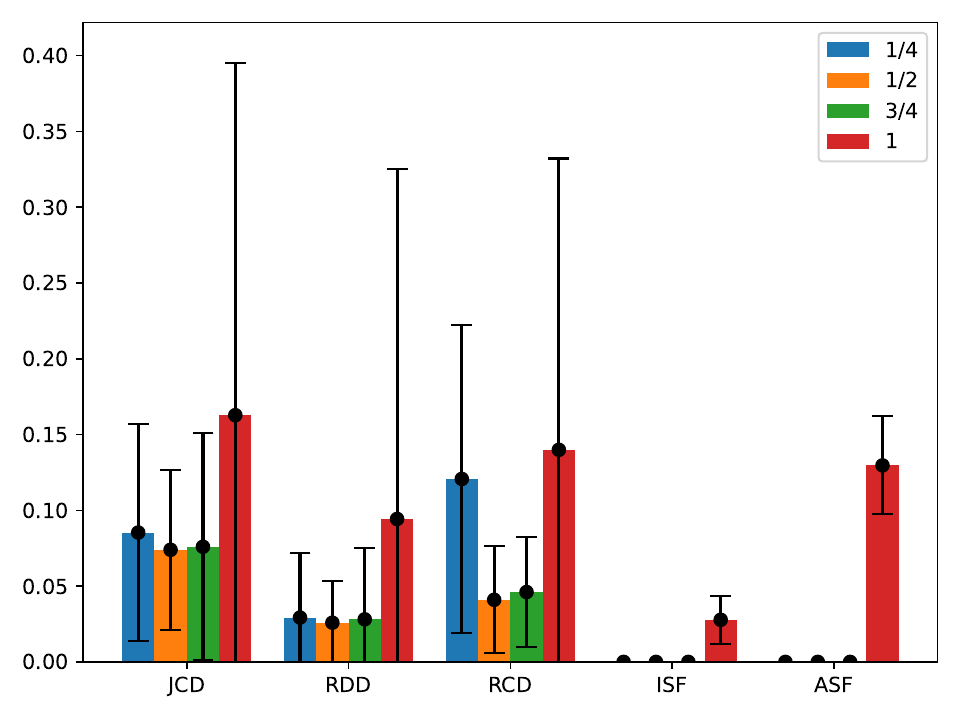}}

    \subfloat{
        \begin{tabular}{|l|r|r|r|r|r|}
            \hline
            Resize factor & \gls{JCD} & \gls{DD} & \gls{CD} & \gls{ISF} & \gls{ASF} \\
            \hline
            \hline
            1/4 & 0.085 & 0.029 & 0.121 & \cellcolor{gray!25}\textbf{0.000} & \cellcolor{gray!25}\textbf{0.000} \\
            $\pm$ & 0.071 & 0.042 & 0.101 & 0.000 & 0.000 \\
            \hline
            1/2 & \cellcolor{gray!25}\textbf{0.074} & \cellcolor{gray!25}\textbf{0.026} & \cellcolor{gray!25}\textbf{0.041} & \cellcolor{gray!25}\textbf{0.000} & \cellcolor{gray!25}\textbf{0.000} \\
            $\pm$ & 0.053 & 0.027 & 0.035 & 0.000 & 0.000 \\
            \hline
            3/4 & 0.076 & 0.028 & 0.046 & \cellcolor{gray!25}\textbf{0.000} & \cellcolor{gray!25}\textbf{0.000} \\
            $\pm$ & 0.075 & 0.047 & 0.036 & 0.000 & 0.000 \\
            \hline
            1 & 0.163 & 0.094 & 0.140 & 0.028 & 0.130 \\
            $\pm$ & 0.232 & 0.231 & 0.192 & 0.016 & 0.032 \\
            \hline
        \end{tabular}
    }
    \caption{Values of evaluation metrics for different image sizes.
      Bold values highlight the optimum in each column. The other
      hyperparameters are fixed at Backbone: ResNet 34, Optimizer:
      Adam, Loss: Cross-Entropy, Transfer learning: Yes, Data
      augmentation: Yes. The U-Net can achieve the highest accuracy
      when the original image size is reduced by half.}
    \label{sifig:resolution_unet}
\end{figure}

\begin{figure}[!ht]
    \centering
    \subfloat{\includegraphics[width=\linewidth]{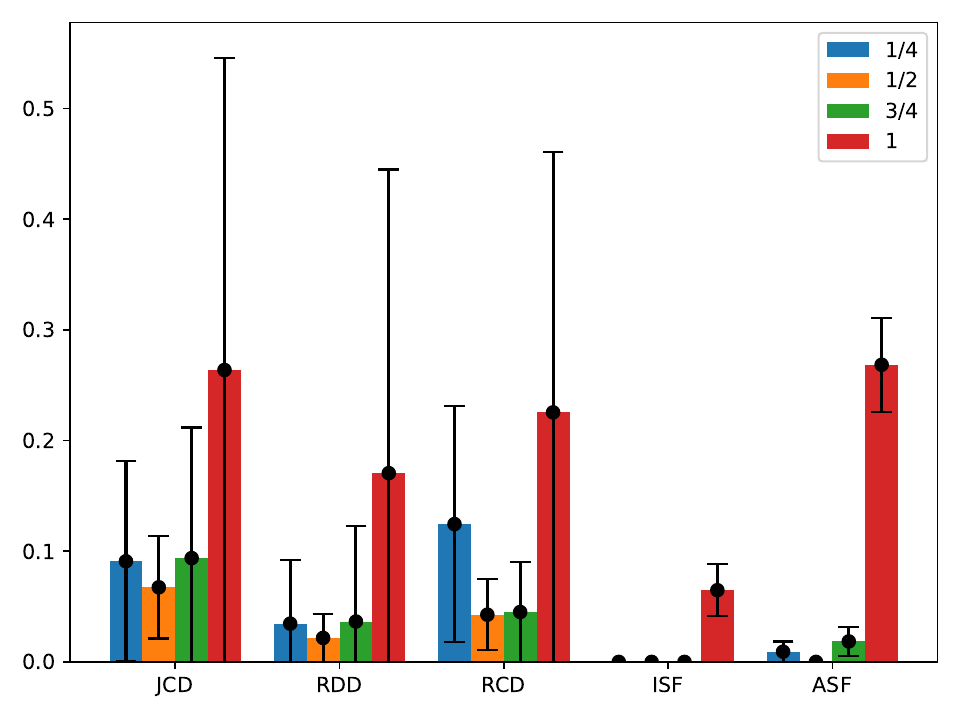}}

    \subfloat{
        \begin{tabular}{|l|r|r|r|r|r|}
            \hline
            Resize factor & \gls{JCD} & \gls{DD} & \gls{CD} & \gls{ISF} & \gls{ASF} \\
            \hline
            \hline
            1/4 & 0.091 & 0.035 & 0.125 & \cellcolor{gray!25}\textbf{0.000} & 0.009 \\
            $\pm$ & 0.090 & 0.058 & 0.106 & 0.000 & 0.009 \\
            \hline
            1/2 & \cellcolor{gray!25}\textbf{0.067} & \cellcolor{gray!25}\textbf{0.022} & \cellcolor{gray!25}\textbf{0.043} & \cellcolor{gray!25}\textbf{0.000} & \cellcolor{gray!25}\textbf{0.000} \\
            $\pm$ & 0.046 & 0.022 & 0.032 & 0.000 & 0.000 \\
            \hline
            3/4 & 0.094 & 0.037 & 0.045 & \cellcolor{gray!25}\textbf{0.000} & 0.019 \\
            $\pm$ & 0.118 & 0.086 & 0.045 & 0.000 & 0.013 \\
            \hline
            1 & 0.264 & 0.171 & 0.226 & 0.065 & 0.269 \\
            $\pm$ & 0.282 & 0.274 & 0.235 & 0.024 & 0.043 \\
            \hline
        \end{tabular}
    }
    \caption{Values of evaluation metrics for different image sizes.
      Bold values highlight the optimum in each column. The other
      hyperparameters are fixed at Backbone: W48, Optimizer: Adam,
      Loss: Cross-Entropy, Transfer learning: Yes, Data augmentation:
      Yes. The HRNet can achieve the highest accuracy when the
      original image size is reduced by half.}
    \label{sifig:resolution_hrnet}
  \end{figure}

  \begin{figure}[!ht]
    \centering
    \subfloat{\includegraphics[width=\linewidth]{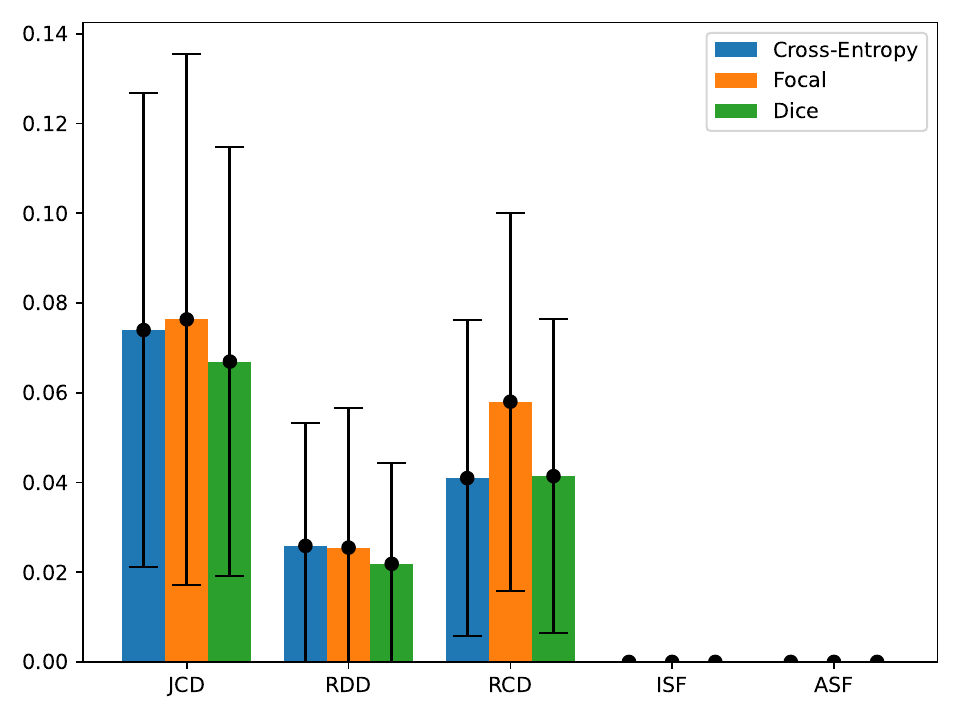}}

    \subfloat{
        \begin{tabular}{|l|r|r|r|r|r|}
            \hline
            Loss function & \gls{JCD} & \gls{DD} & \gls{CD} & \gls{ISF} & \gls{ASF} \\
            \hline
            \hline
            Cross-Entropy & 0.074 & 0.026 & 0.041 & \cellcolor{gray!25}\textbf{0.000} & \cellcolor{gray!25}\textbf{0.000} \\
            $\pm$ & 0.053 & 0.027 & 0.035 & 0.000 & 0.000 \\
            \hline
            Focal & 0.076 & 0.025 & 0.058 & \cellcolor{gray!25}\textbf{0.000} & \cellcolor{gray!25}\textbf{0.000} \\
            $\pm$ & 0.059 & 0.031 & 0.042 & 0.000 & 0.000 \\
            \hline
            Dice & \cellcolor{gray!25}\textbf{0.067} & \cellcolor{gray!25}\textbf{0.022} & \cellcolor{gray!25}\textbf{0.041} & \cellcolor{gray!25}\textbf{0.000} & \cellcolor{gray!25}\textbf{0.000} \\
            $\pm$ & 0.048 & 0.023 & 0.035 & 0.000 & 0.000 \\
            \hline
        \end{tabular}
    }
    \caption{Values of evaluation metrics for different loss
      functions, used by the U-Net. Bold values highlight the optimum
      in each column. The other hyperparameters are fixed at Backbone:
      ResNet 34, Optimizer: Adam, Resize factor: 1/2, Transfer
      learning: Yes, Data augmentation: Yes. The U-Net can achieve the
      highest accuracy when the Dice loss is used.}
    \label{sifig:loss_unet}
\end{figure}

\begin{figure}[!ht]
    \centering
    \subfloat{\includegraphics[width=\linewidth]{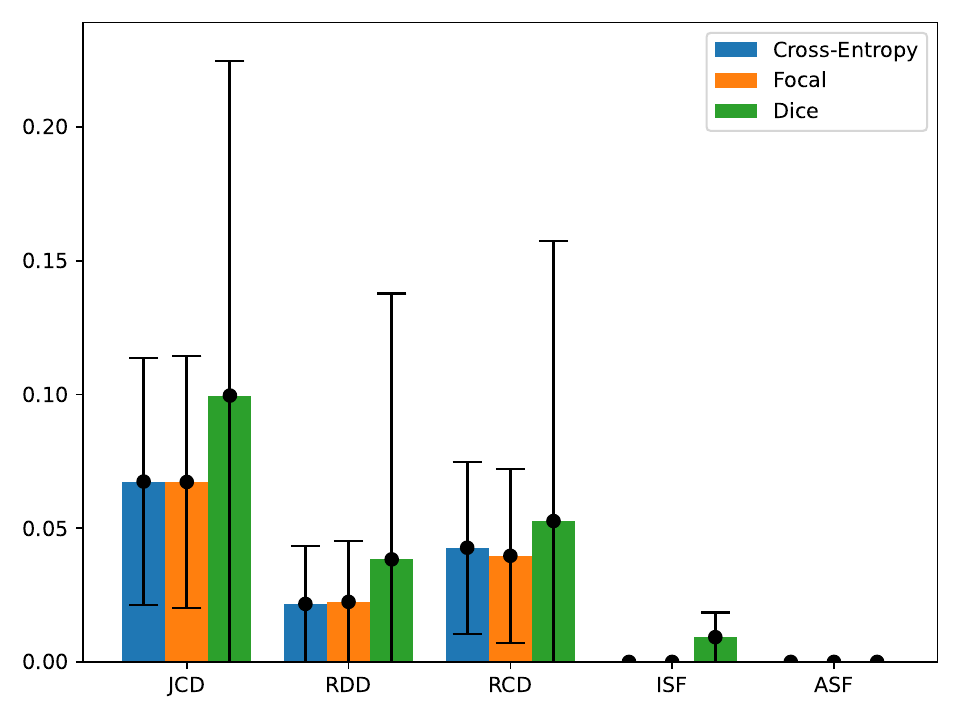}}

    \subfloat{
        \begin{tabular}{|l|r|r|r|r|r|}
            \hline
            Loss function & \gls{JCD} & \gls{DD} & \gls{CD} & \gls{ISF} & \gls{ASF} \\
            \hline
            \hline
            Cross-Entropy & \cellcolor{gray!25}\textbf{0.067} & \cellcolor{gray!25}\textbf{0.022} & 0.043 & \cellcolor{gray!25}\textbf{0.000} & \cellcolor{gray!25}\textbf{0.000} \\
            $\pm$ & 0.046 & 0.022 & 0.032 & 0.000 & 0.000 \\
            \hline
            Focal & \cellcolor{gray!25}\textbf{0.067} & \cellcolor{gray!25}\textbf{0.022} & \cellcolor{gray!25}\textbf{0.040} & \cellcolor{gray!25}\textbf{0.000} & \cellcolor{gray!25}\textbf{0.000} \\
            $\pm$ & 0.047 & 0.023 & 0.032 & 0.000 & 0.000 \\
            \hline
            Dice & 0.100 & 0.038 & 0.053 & 0.009 & \cellcolor{gray!25}\textbf{0.000} \\
            $\pm$ & 0.125 & 0.099 & 0.105 & 0.009 & 0.000 \\
            \hline
        \end{tabular}
    }
    \caption{Values of evaluation metrics for different loss
      functions, used by the HRNet. Bold values highlight the optimum
      in each column. The other hyperparameters are fixed at Backbone:
      W48, Optimizer: Adam, Resize factor: 1/2, Transfer learning:
      Yes, Data augmentation: Yes. The HRNet can achieve the highest
      accuracy, if the Cross-Entropy or the Focal loss is used.}
    \label{sifig:loss_hrnet}
  \end{figure}

  \begin{figure}[!ht]
    \centering
    \subfloat{\includegraphics[width=\linewidth]{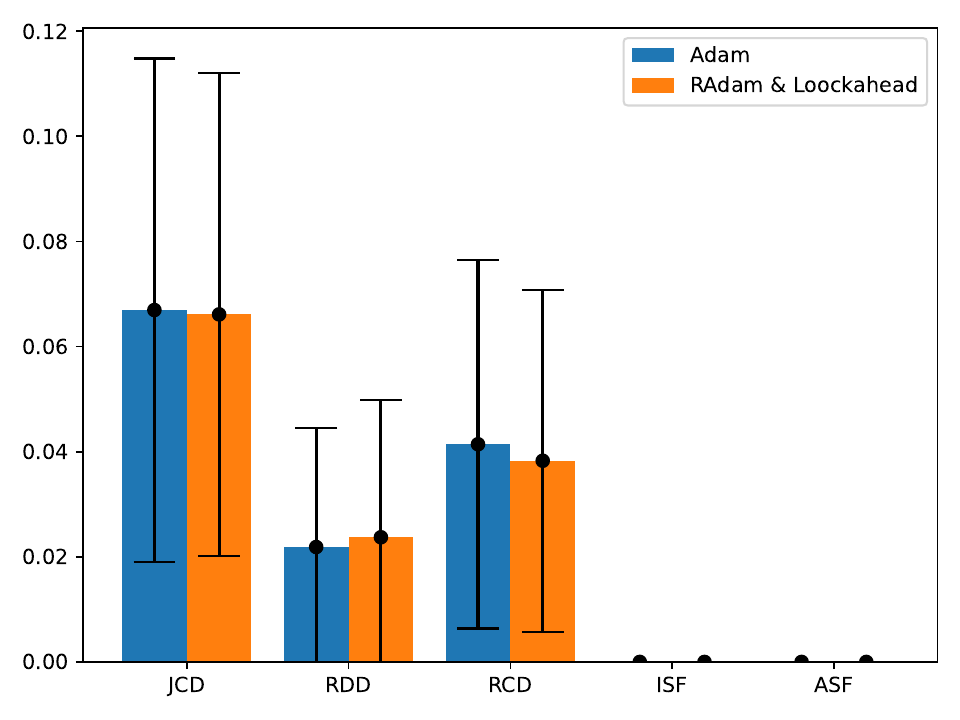}}

    \subfloat{
        \begin{tabular}{|l|r|r|r|r|r|}
            \hline
            Optimizer function & \gls{JCD} & \gls{DD} & \gls{CD} & \gls{ISF} & \gls{ASF} \\
            \hline
            \hline
            Adam & 0.067 & \cellcolor{gray!25}\textbf{0.022} & 0.041 & \cellcolor{gray!25}\textbf{0.000} & \cellcolor{gray!25}\textbf{0.000} \\
            $\pm$ & 0.048 & 0.023 & 0.035 & 0.000 & 0.000 \\
            \hline
            RAdam \& Lookahead & \cellcolor{gray!25}\textbf{0.066} & 0.024 & \cellcolor{gray!25}\textbf{0.038} & \cellcolor{gray!25}\textbf{0.000} & \cellcolor{gray!25}\textbf{0.000} \\
            $\pm$ & 0.046 & 0.026 & 0.033 & 0.000 & 0.000 \\
            \hline
        \end{tabular}
    }
    \caption{Values of evaluation metrics for different optimizers,
      used by the U-Net. Bold values highlight the optimum in each
      column. The other hyperparameters are fixed at Backbone: ResNet
      34, Loss: Dice, Resize factor: 1/2, Transfer learning: Yes, Data
      augmentation: Yes. The U-Net can achieve the highest accuracy
      when the optimization is done by RAdam combined with Lookahead.}
    \label{sifig:optimizer_unet}
\end{figure}

\begin{figure}[!ht]
    \centering
    \subfloat{\includegraphics[width=\linewidth]{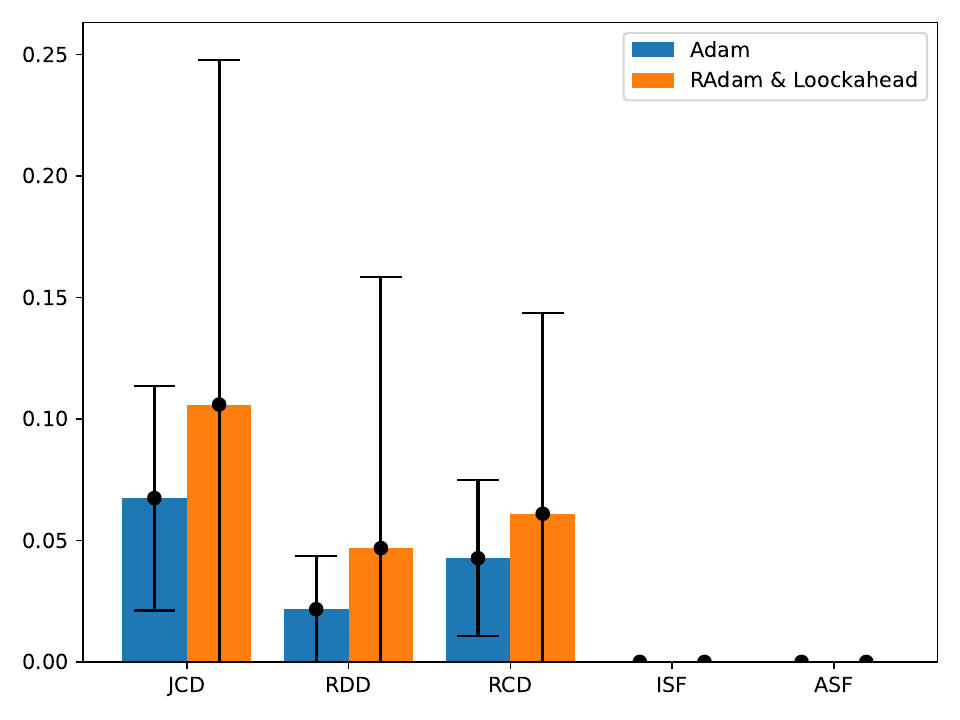}}

    \subfloat{
        \begin{tabular}{|l|r|r|r|r|r|}
            \hline
            Optimizer function & \gls{JCD} & \gls{DD} & \gls{CD} & \gls{ISF} & \gls{ASF} \\
            \hline
            \hline
            Adam & \cellcolor{gray!25}\textbf{0.067} & \cellcolor{gray!25}\textbf{0.022} & \cellcolor{gray!25}\textbf{0.043} & \cellcolor{gray!25}\textbf{0.000} & \cellcolor{gray!25}\textbf{0.000} \\
            $\pm$ & 0.046 & 0.022 & 0.032 & 0.000 & 0.000 \\
            \hline
            RAdam \& Lookahead & 0.106 & 0.047 & 0.061 & \cellcolor{gray!25}\textbf{0.000} & \cellcolor{gray!25}\textbf{0.000} \\
            $\pm$ & 0.142 & 0.112 & 0.083 & 0.000 & 0.000 \\
            \hline
        \end{tabular}
    }
    \caption{Values of evaluation metrics for different optimizers,
      used by the HRNet. Bold values highlight the optimum in each
      column. The other hyperparameters are fixed at Backbone: W48,
      Loss: Cross-Entropy, Resize factor: 1/2, Transfer learning: Yes,
      Data augmentation: Yes. The HRNet can achieve the highest
      accuracy when the optimization is done by Adam.}
    \label{sifig:optimizer_hrnet}
  \end{figure}

 \begin{figure}[!ht]
    \centering
    \subfloat{\includegraphics[width=\linewidth]{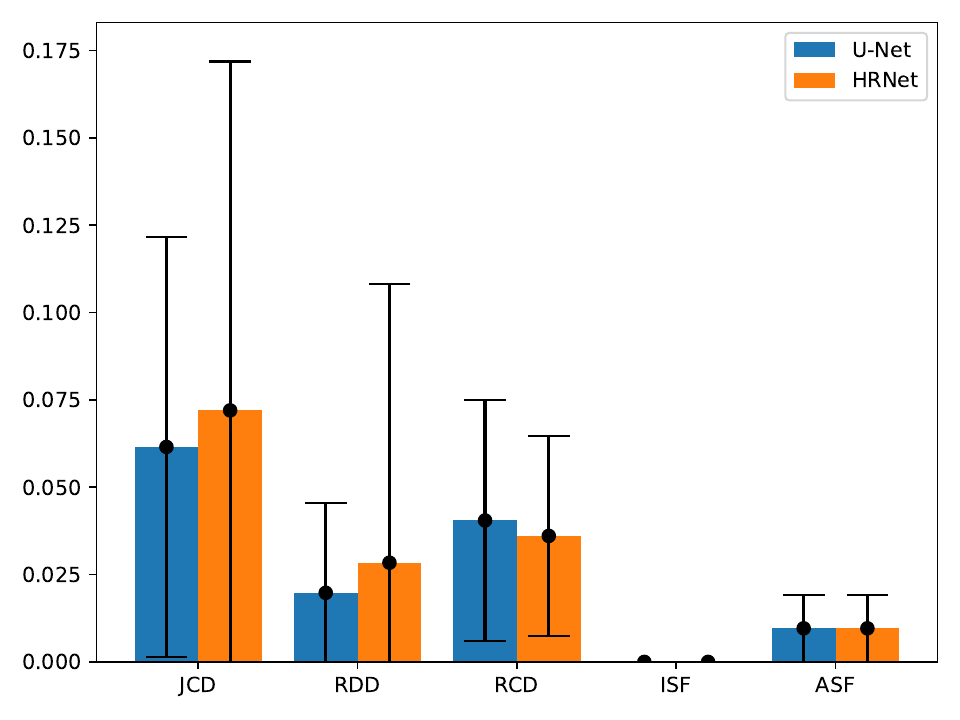}}

    \subfloat{
        \begin{tabular}{|l|r|r|r|r|r|}
            \hline
            Model & \gls{JCD} & \gls{DD} & \gls{CD} & \gls{ISF} & \gls{ASF} \\
            \hline
            \hline
            U-Net & \cellcolor{gray!25}\textbf{0.062} & \cellcolor{gray!25}\textbf{0.020} & 0.040 & \cellcolor{gray!25}\textbf{0.000} & \cellcolor{gray!25}\textbf{0.010} \\
            $\pm$ & 0.060 & 0.026 & 0.034 & 0.000 & 0.010 \\
            \hline
            HRNet & 0.072 & 0.028 & \cellcolor{gray!25}\textbf{0.036} & \cellcolor{gray!25}\textbf{0.000} & \cellcolor{gray!25}\textbf{0.010} \\
            $\pm$ & 0.100 & 0.080 & 0.029 & 0.000 & 0.010 \\
            \hline
        \end{tabular}
    }
    \caption{Evaluation of the segmentation with the optimized U-Net
      and HRNet models on the test data set shows higher accuracy of
      the U-Net. The optimal hyperparameter configuration for the
      U-Net is Backbone: ResNet 34, Optimizer: RAdam \& Lookahead,
      Loss: Dice, Resize factor: 1/2, Transfer learning: Yes, Data
      augmentation: Yes. The final hyperparameter configuration for
      the HRNet is Backbone: W48, Optimizer: Adam, Loss:
      Cross-Entropy, Resize factor: 1/2, Transfer learning: Yes, Data
      augmentation: Yes.}
    \label{sifig:SItestset}
  \end{figure}

    \begin{figure*}[!ht]
    \centering
    \subfloat[Image with segmentations]{\includegraphics[width=0.33\linewidth,trim=7cm 9cm 8cm 0cm,clip]{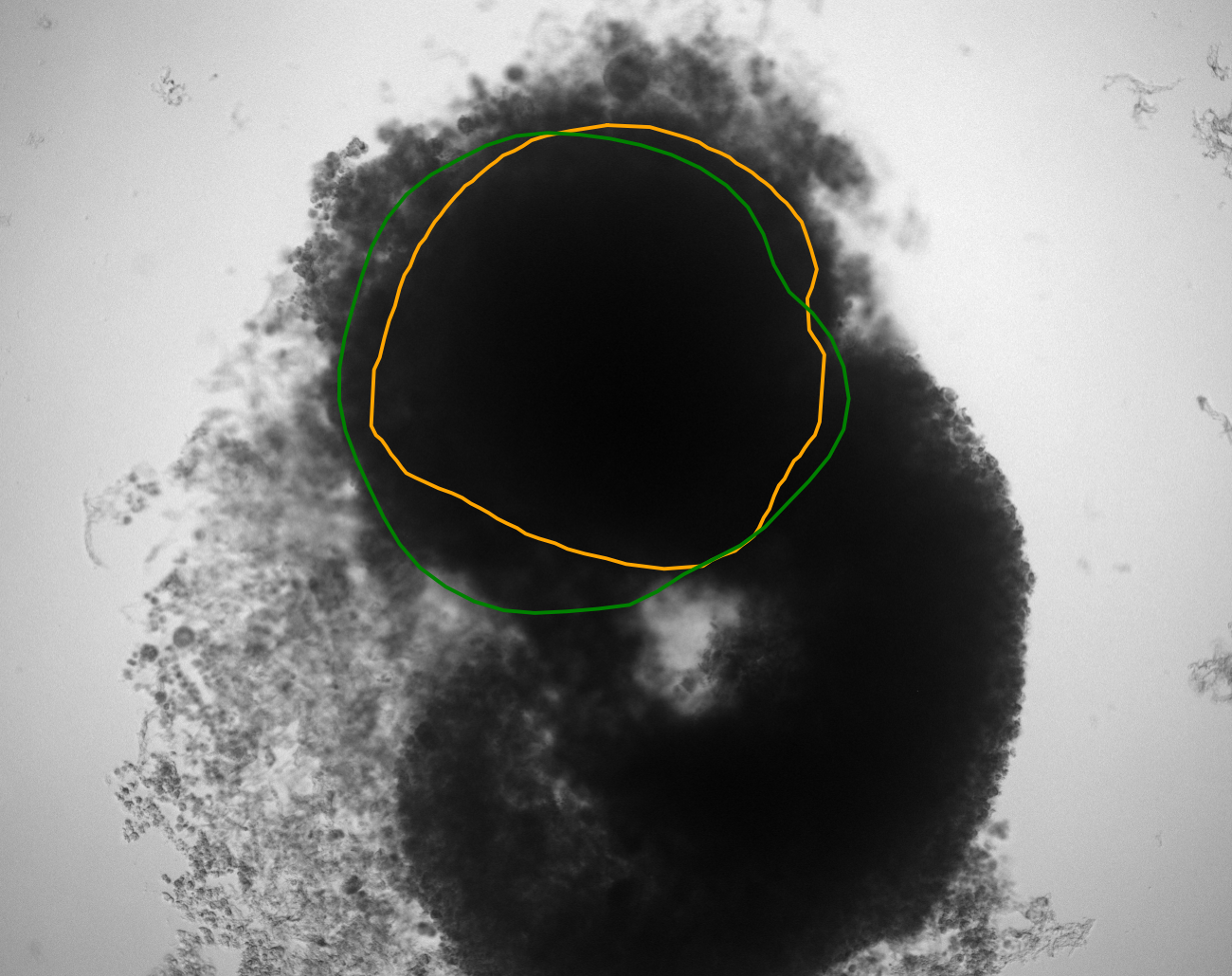}}
    \subfloat[Segmentations with mismatched (gray) areas]{\includegraphics[width=0.33\linewidth,trim=7cm 9cm 8cm 0cm,clip]{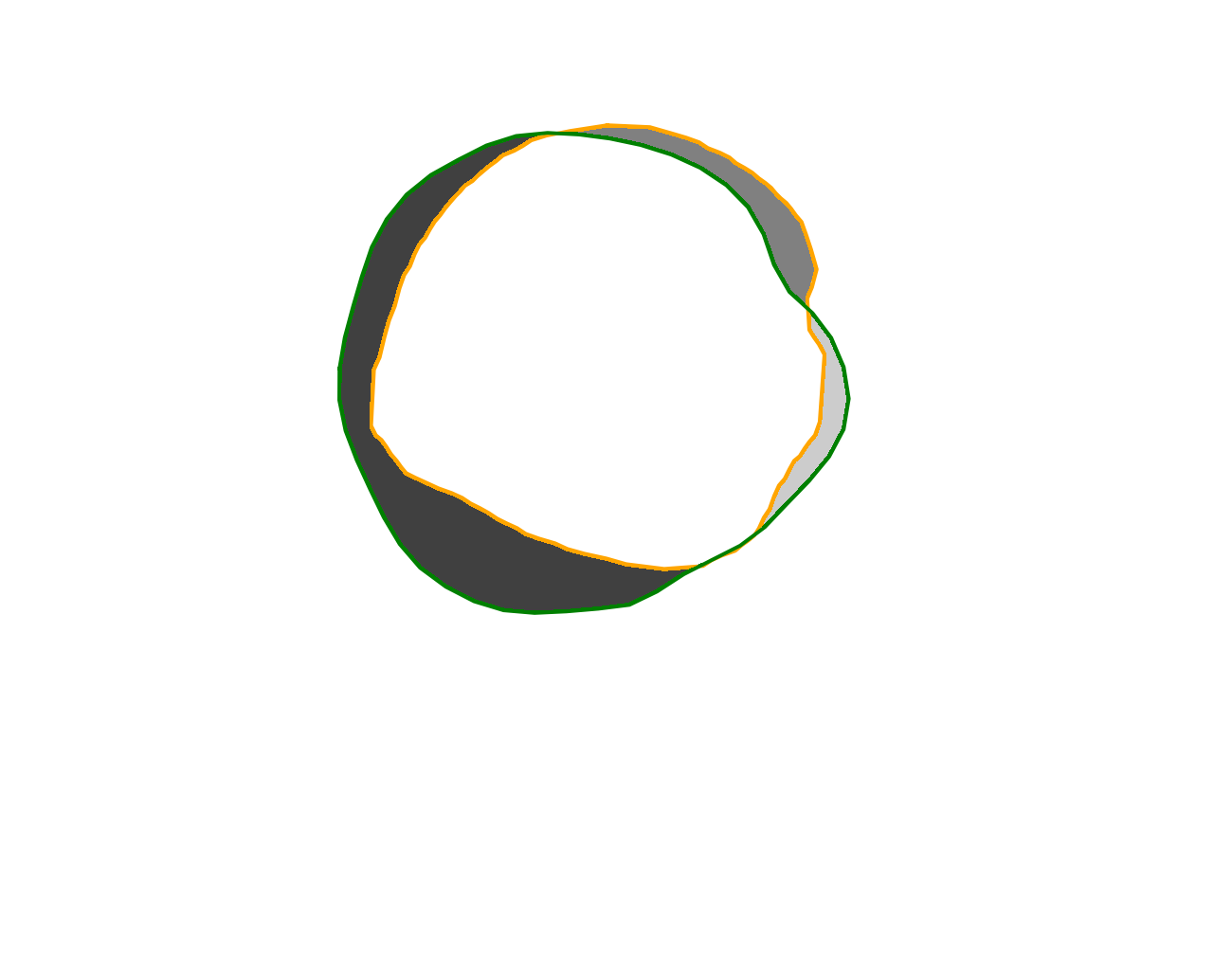}}
    \subfloat[Corresponding circle with rim of width $\Delta r$]{\includegraphics[width=0.33\linewidth,trim=7cm 5cm 8cm 5cm,clip]{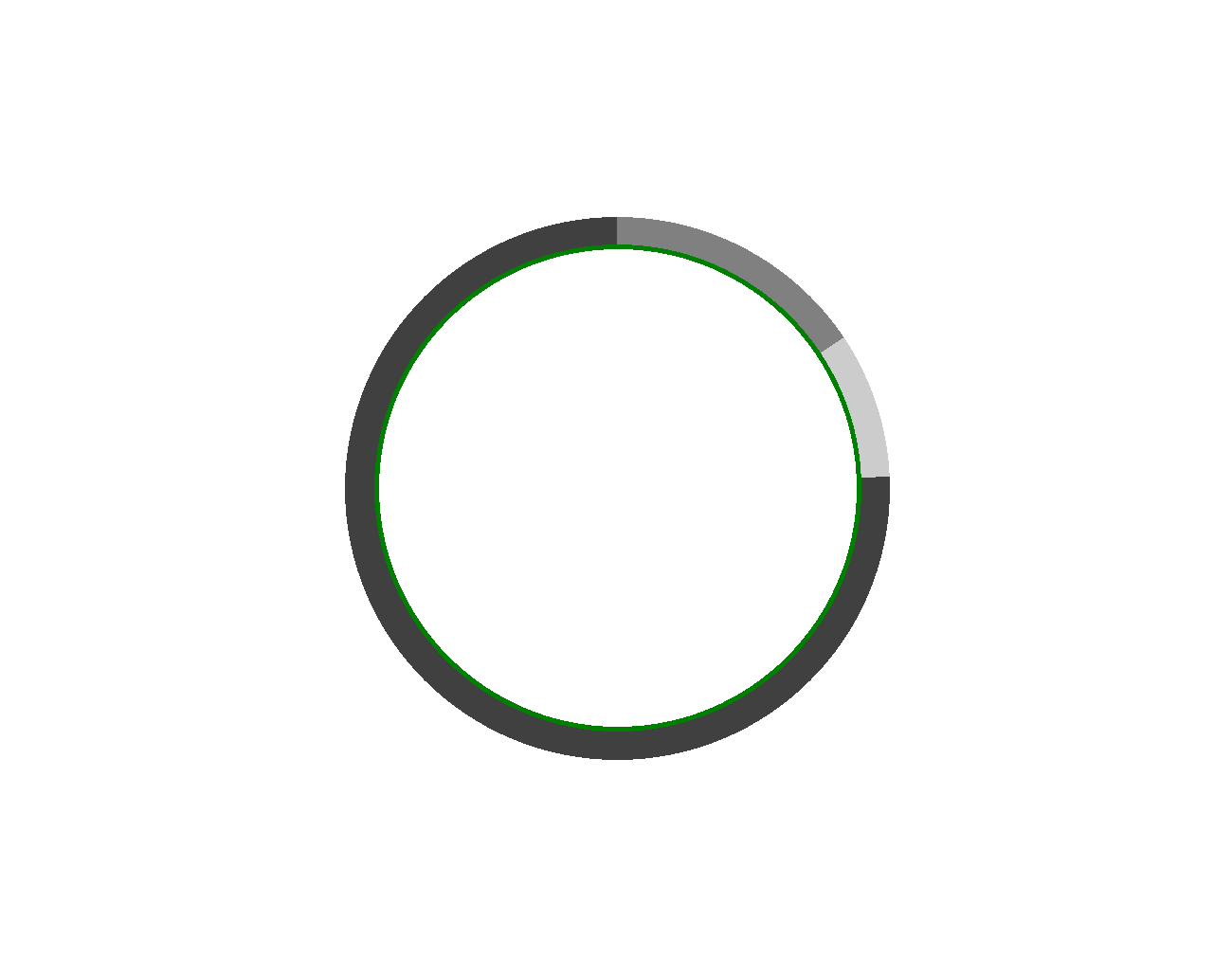}}
    \caption{\label{sifig:ARE} Illustration of the average radial
      error $\Delta r$ defined in Eq.~(\ref{eq:dr}) (example with
      large $\gls{JCD} > 0.2$ for visibility). (a) Zoom of image from
      \prettyref{sifig:samplesJCD30} with $\gls{JCD} = 0.24$ with
      corresponding automatic (orange line containing predicted area
      $P$) and manual segmentation (green line containing target area
      $T$). (b) Segmentations from (a) with mismatched areas shown in
      gray (missing area $T\setminus P$ in bright and dark gray,
      additional area $P\setminus T$ in intermediate gray). (c) Circle
      (green line) and added rim (gray areas) with areas corresponding
      to (b): Area within green circle is equal to area within manual
      segmentation in (b) and gray areas within the adjacent rim are
      equal to corresponding mismatched areas in (b). Then the radial
      thickness of the rim is equal to the average radial error
      $\Delta r$ defined in Eq.~(\ref{eq:dr}).}
    \end{figure*}

  \begin{figure}[!ht]
    \centering
    \subfloat[\gls{DD} over spheroid
    diameter]{\hspace*{-0.5cm}\includegraphics{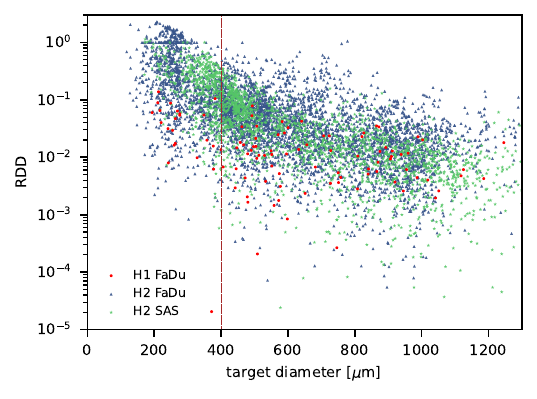}}

    \subfloat[Comparison diameters from manual and automatic segmentation]{\includegraphics{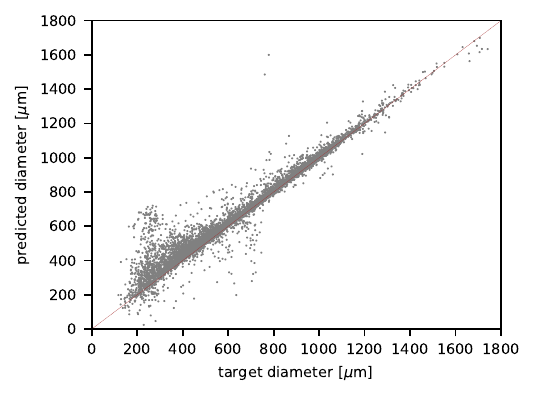}}
    \caption{\label{sifig:RDD} Validation based on diameter of
      automatic segmentation with the optimized U-Net on larger,
      independent data sets, analogous to \prettyref{fig:validation}.
      (a) \gls{DD} shows that majority of deviations are below $10 \%$
      and higher deviations occur mostly below $d_T<400 \ \mu$m. (b)
      Direct comparison of diameters (gray points) resulting from
      automatic segmentation (predicted diameter) and manual
      segmentation (target diameter) shows high accuracy (red line
      represents perfect match) with larger deviations mostly at
      $d_T<400 \ \mu$m.}
    \end{figure}

\end{document}